\documentclass[conference,12pt,onecolumn,draftclsnofoot,]{IEEEtran}

\usepackage{subfigure}
\ifCLASSINFOpdf
   \usepackage[pdftex]{graphicx}
  \graphicspath{{../pdf/}{../jpeg/}}
   \DeclareGraphicsExtensions{.pdf,.jpeg,.png}
\else
\fi

\usepackage{times}
\usepackage{subfigure}
\usepackage{amsfonts}
\usepackage{amssymb}
\usepackage{dsfont}
\usepackage{amsmath}
\usepackage{epsfig}
\usepackage{xspace}
\usepackage{graphicx}
\usepackage{multirow}
\usepackage{epstopdf}
\usepackage{amssymb}
\usepackage{lineno,hyperref}
\modulolinenumbers[5]

\usepackage[noend]{algorithmic} 
\usepackage{algorithm,caption}
\algsetup{indent=2em}

\makeatletter
\def\BState{\State\hskip-\ALG@thistlm}
\makeatother

\begin{document}
%
\title{Hashing in the Zero Shot Framework with Domain Adaptation}


\author{\IEEEauthorblockN{Shubham Pachori\IEEEauthorrefmark{1},
Ameya Deshpande\IEEEauthorrefmark{2} and Shanmuganathan Raman\IEEEauthorrefmark{3}}
\IEEEauthorblockA{Electrical Engineering,
Indian Institute of Technology Gandhinagar\\
Gujarat, India 382355\\
Email: $\lbrace$ \IEEEauthorrefmark{1}shubham\textunderscore pachori,
\IEEEauthorrefmark{2}deshpande.ameya,
\IEEEauthorrefmark{3}shanmuga $\rbrace$ @iitgn.ac.in}}


%


\maketitle

\begin{abstract}
Techniques to learn hash codes which can store and retrieve large dimensional multimedia data efficiently have attracted broad research interests in the recent years. With rapid explosion of newly emerged concepts and online data, existing supervised hashing algorithms suffer from the problem of scarcity of ground truth annotations due to the high cost of obtaining manual annotations. Therefore, we propose an algorithm to learn a hash function from training images belonging to `seen' classes  which can efficiently encode images of `unseen' classes to binary codes. Specifically, we project the image features from visual space and semantic features from semantic space into a common Hamming subspace. Earlier works to generate hash codes have tried to relax the discrete constraints on hash codes and solve the continuous optimization problem. However, it often leads to quantization errors. In this work, we use the max-margin classifier to learn an efficient hash function. To address the concern of domain-shift which may arise due to the introduction of new classes, we also introduce an unsupervised domain adaptation model in the proposed hashing framework. Results on the three datasets show the advantage of using domain adaptation in learning a high-quality hash function and superiority of our method for the task of image retrieval performance as compared to several state-of-the-art hashing methods. 
\end{abstract}
%
\IEEEpeerreviewmaketitle

\section{Introduction}
Hashing based techniques \cite{wang2014hashing} are widely used to encode a high dimensional data in a larger scale in a compact binary codes (i,e. 0 or 1). In this way, large scale multimedia data could be stored compactly and could be retrieved efficiently by calculating the Hamming distance between binary codes in a shorter response time.
Because of its efficacy in dealing with the problems of `curse of dimensionality', it has been used in many computer vision applications, including feature points matching \cite{strecha2012ldahash}, multimedia event detection \cite{petrovic2010streaming}, and video segmentation \cite{liu2014weakly}.
With the rapidly increasing amount of multimedia web data, standard hashing methods face tremendous challenges in leveraging reliable ground truth for the available data due to the time-consuming process of manual annotations. Humans and machines still acquire visual concepts in different ways. 
Machines, similar to child, try to mimic the process of learning by visual examples to take decisions based on previously learned patterns.
As our understanding grows, a lot of our diverse knowledge comes from information from other modalities (particularly text and oral). 
Thus, by exploiting the semantic correlations in different modalities, it is possible for human intelligence to identify an object of completely new category by transferring knowledge acquired from known categories.
Zero shot learning methods bridge the semantic gap between `seen' and `unseen' categories by transferring supervised knowledge (using class-attribute descriptors) between `seen' and `unseen' categories. By leveraging information from other modalities for example by learning word embeddings that capture distributional similarity in the textual domain \cite{turian2010word} from large scale text corpus such as Wikipedia, we can ground visual modality and transfer such knowledge into an optimization model. In the hashing domain, However, the zero-shot problem has rarely been studied. As per best of our knowledge, only two works \cite{yang2016zero} and \cite{pachori2016zero} have addressed this problem previously. In Section 2, we briefly review some related work on hashing and zero-shot learning. In Section 3, we elaborate our approach with details, together with our optimization method and an analysis of the algorithm. With extensive experiments, various results on various different datasets are reported in Section 4, followed by the conclusion of the work in Section 5.

\section{Related Works}
Similarity search is one of the key challenges in machine learning applications.
Most of the existing hashing methods generate binary codes by exploiting pairwise class label information to efficiently find similar data examples to a query. 
Broadly, hashing schemes can be categorized as either data-independent hashing or data-dependent hashing.
No supervised knowledge or prior information is assumed for data-independent hashing techniques.
Locality Sensitive Hashing \cite{gionis1999similarity} is one of the most popular examples in data-dependent category.
In general, data-independent hashing schemes generate hash functions randomly and require large number of hash look up tables to achieve reasonable performance. To tackle this issue, in the last decade, researchers have focussed strongly to improve data-dependent hashing methods. 
Further data-dependent hashing schemes can be categorized into supervised or unsupervised learning methods.
Supervised hashing methods (example Supervised Discrete Hashing \cite{shen2015supervised}, Column Sampling Discrete Hashing \cite{kang2016column}) generally achieve better performance than unsupervised learning based hashing techniques (example Iterative Quantization \cite{gong2011iterative} and Inductive Hashing on Manifolds \cite{shen2013inductive}). Unsupervised hashing techniques, in general, use manifold learning techniques or graph cut techniques to extract the intrinsic structure of the data embedded in the feature space. Algorithms have also been proposed to encode the mid-level information like attributes present in the image in its hash code \cite{li2015two}. Given that almost all the data available on internet is recounted in the multiple modalities, like semantic information (for example, image captions) and visual features. Because of availability and incorporating supervised information into learning model, supervised hashing schemes, in general, perform better than unsupervised ones. however, what if we want to achieve data-dependent performance while no training example is provided? 
All the mentioned hashing methods are limited to only seen categories i.e. at least one example correspond to each category is present in training set but would fail to generalize to  any ``unseen" category. 

Zero shot learning (\cite{lampert2014attribute},\cite{akata2013label},\cite{romera2015embarrassingly},\cite{xian2016latent})  have proved its efficacy in learning concepts with no available training examples, thereby tackling the problem posed by increasing large amount of online data with no ground truth annotations. Because training and testing categories are disjoint, one cannot directly apply the supervised learning algorithms with per-image class labels. Zero shot learning algorithms assume that there is a space within which both visual and semantic features could be embedded simultaneously. Most commonly adopted semantic embedding space is an attribute space where the class labels or categories are represented by a vector which quantifies the amount of each attribute generally found in the instances or examples of that class \cite{farhadi2009describing}. But an attribute ontology has to be manually defined to describe a class using an attribute vector. To address this issue, methods have been developed to leverage large textual corpus such as Wikipedia to obtain semantic features for class labels by exploiting the correlations between different words. \cite{socher2013zero} and  \cite{frome2013devise} are examples of zero shot learning works which obtain semantic features from NLP techniques.

Our main idea is motivated from structured embedding frameworks (\cite{akata2013label}, \cite{romera2015embarrassingly}). We first represent both the images' features and seen classes' features in a common multi-dimensional Hamming subspace. Image features can be obtained from state-of-the-art image representations, for example from gist features \cite{oliva2001modeling}. Class features, as discussed above can either (i) be extracted automatically (\cite{mikolov2013distributed}, \cite{pennington2014glove}) from an unlabelled large text corpora (ii) or can be obtained using manually specified side information example attributes \cite{lampert2014attribute}. In the learning procedure, we also incorporate an objective function that learns to pull images from the same class and having similar visual representation in visual space close to each other. But learning a hash function from a naive knowledge transfer using source domain without making it adapt to the target domain may lead to severe domain shift problems. To deal with this issue, we also incorporate the domain adaptation technique in hash code learning procedure. Once learned, the hash function can be used to generate the hash codes for any seen as well as unseen category images and thus enabling zero-shot hashing which can at the same time, address the domain shift problem.

\section{Methodology}
In section 3.1, we introduce the notations used throughout this paper. In section 3.2, we introduce the objective function to embed the attribute/textual information of the classes and representation of images into a common Hamming space. In section 3.3, we introduce a measure to preserve the intra-modal similarity information in the hash codes. In sections 3.4 and 3.5, we introduce an efficient optimization scheme for the proposed objective function to learn the hash function. In section 3.6, we introduced domain adaptaion in the hashing framework.

\subsection{Notation}

Let the number of training images in `seen' classes be $N$, each of which is represented by $d_x$ dimension features. Feature vector of $i$th image is represented by $\boldsymbol{x_{i}}$. An image data matrix $\boldsymbol{X} \in \mathbb{R}^{(d_{x} \times N)}$ is formed by arranging these image features column wise, that is, $i$th column of $\boldsymbol{X}$ contains $\boldsymbol{x_{i}}$.
Let the number of `seen' and `unseen' classes be $n_{s}$ and $n_{u}$ respectively. Moreover, the ``unseen" and ``seen" classes are disjoint (as per the hypothesis of zero shot learning approaches \cite{romera2015embarrassingly} \cite{kodirov2015unsupervised}).
Each of the $n_{s} + n_{u}$ classes is represented by the respective attributes/textual information vector. 
For each image, the attribute/textual information is stored in vector of length $d_y$. Like image data matrix $\boldsymbol{X}$, textual data matrix $\boldsymbol{Y} \in \mathbb{R}^{ (d_{y}\times N)}$ is generated by arranging attribute/textual information of all images column wise.
Each value in $\boldsymbol{Y}$ represents how strongly a particular attribute is evident in a given image.
Let the length of the hash codes be $K$.
Our objective is to learn binary codes $\boldsymbol{B} \in \lbrace -1, 1\rbrace^{(N \times K)}$ for all the $N$ images of the seen categories as well as to learn a hash function to generate hash codes for new images and images belonging to `unseen' categories. 
Without any loss of generality, we consider to learn a linear form of hash function $sgn(\boldsymbol{x_{i}}^{\intercal} \boldsymbol{W_{img}})$ to generate hash codes using image features $\boldsymbol{x_{i}}$, where $\boldsymbol{W_{img}} \in \mathbb{R}^{(d_{x}\times K)}$. 
To embed the information of textual feature data into a common Hamming space, a transformation matrix $\boldsymbol{W_{txt}} \in \mathbb{R}^{(d_{y}\times K)}$ is learned through the optimization scheme proposed in section 3.4.

\subsection{Inter-modality Preservation}
\begin{equation} \label{firstobjectivefunction}
\underset{\boldsymbol{W_{img}},\boldsymbol{W_{txt}}}{\mathrm{\textit{arg} min}} ||(\boldsymbol{Y}^\intercal \boldsymbol{W_{txt}}) - (sgn(\boldsymbol{X}^\intercal \boldsymbol{W_{img}}))||^{2}_{2} + \beta ||\boldsymbol{W_{txt}}||_{2}^{2}\\
\end{equation}
Eq. \ref{firstobjectivefunction} represents the objective function in order to minimize the squared inter-modal loss. $W_{txt}$ is a transformation matrix which embeds the semantic features into Hamming Space. Though, we do not restrict $\boldsymbol{Y}^\intercal \boldsymbol{W_{txt}} \in \lbrace -1, +1 \rbrace^{N \times K}$ but to take values as much as possible to $\lbrace -1, +1 \rbrace^{N \times K}$. Therefore, we do not impose the orthogonality constraint on $\boldsymbol{W_{txt}}$. 


\subsection{Intra-Modal Similarity Preservation}

We exploit the local structural information in visual feature space to preserve the intra-modal similarity between data points i.e., the hash codes of similar images should have less Hamming distance. To preserve this intra-modal similarity, we ensure that the feature points present nearby in the higher dimension $\boldsymbol{x_{i}}$ must be present close to each other in lower dimension (i.e. in Hamming space) as well, which preserves \textit{Feature-to-Feature Structure Similarity}. Moreover, the hash codes of the images belonging to the same category should be similar in order to preserve \textit{Class-to-Class Information Similarity}.

\subsubsection{Feature-to-Feature Structure Similarity Preservation}
For similar (dissimilar) pairs, distance between them is expected to be minimized (maximized) in the Hamming space. Higher dimensional features having similar representations should also lie close to each other in the lower dimensional space.
This reflects the manifold assumption that visually similar images are more likely to embed closer in the Hamming space. Specifically, if $\boldsymbol{x_{p}}$ and $\boldsymbol{x_{q}}$ lie near to each other in the higher dimensional space, they should share similar hash codes. To incorporate the neighborhood similarity in hash codes, we adopt the $k$ nearest neighbourhood function as given in \cite{weiss2009spectral} in order to calculate the affinity matrix $\boldsymbol{W^{F}}$ as shown in Eq. \ref{featuretofeatureimilarity}:
\begin{equation} \label{featuretofeatureimilarity}
  {W^{F}_{pq}} = \left \{
  \begin{aligned}
    &  exp\Bigg(\frac{-||\boldsymbol{x_{p}} - \boldsymbol{x_{q}}||^{2}_{2}}{\sigma^{2}}\Bigg), && \text{if}\ \boldsymbol{x_{p}} \in \mathbb{N}_{k}(\boldsymbol{x_{q}})\hspace{0.1cm}  \text{or}\hspace{0.1cm} \boldsymbol{x_{q}} \in \mathbb{N}_{k}\hspace{0.05cm}(\boldsymbol{x_{p}}) \hspace{0.1cm} \\
    & \hspace{1.5cm}0, && \text{otherwise}
  \end{aligned} \right.
\end{equation}
where $\boldsymbol{p},\boldsymbol{q}=1,2,\ldots,N$ and $\mathbb{N}_{k}$ is the $k$ nearest neighbour set for a given image.

\subsubsection{Class-to-Class Similarity Preservation}
To preserve the class discriminability of binary codes in Hamming space, we decompose the discriminability constraint into two components, i.e., intra-category compactness and inter-category separability.
Both these constraints enforce the lesser Hamming distance between the hash codes of images belonging to the similar categories. To incorporate the intra class similarity information, we define $\boldsymbol{W^{intra}}$ as shown in Eq. \ref{intraclasssimilarity} 
\begin{equation} \label{intraclasssimilarity}
  {W^{intra}_{pq}} = \left \{
  \begin{aligned}
    &  \hspace{0.4cm}1, \hspace{1cm} \text{if}\ \boldsymbol{x}_{p}\hspace{0.1cm} \text{and}\ \boldsymbol{x}_{q}\hspace{0.1cm} \text{belong to the same class}\hspace{0.1cm}   \\
    & \hspace{0.4cm} 0, \hspace{0.96cm}  \text{otherwise}
  \end{aligned} \right.
\end{equation}
where $\boldsymbol{p},\boldsymbol{q}=1,2,\ldots,N$.\\
To incorporate the inter class similarity information, we define $\boldsymbol{W^{inter}}$ as shown in Eq. \ref{interclasssimilarity} 
\begin{equation} \label{interclasssimilarity}
  {W^{inter}_{pq}} = \left \{
  \begin{aligned}
    &  \hspace{0.4cm}1, \hspace{1cm} \text{if}\ \boldsymbol{x}_{p}\hspace{0.1cm} \text{and}\ \boldsymbol{x}_{q}\hspace{0.1cm} \text{belong to different classes}\hspace{0.1cm}   \\
    & \hspace{0.4cm} 0, \hspace{0.96cm}  \text{otherwise}
  \end{aligned} \right.
\end{equation}
where $\boldsymbol{p},\boldsymbol{q}=1,2,\ldots,N$.\\
By minimizing the energy function given in Eq. \ref{similaritynew}, we preserve both the feature-to-feature structure information and within class similarity information.
\begin{equation}\label{similaritynew}
\sum_{(i,j)} \frac{{W^{S}_{ij}}}{4K}||\boldsymbol{b_{i}}^\intercal - \boldsymbol{b_{j}}^\intercal ||^{2}_{2}
\end{equation} 
where $ \boldsymbol{W^{S}} = \boldsymbol{W^{intra}} + \boldsymbol{W^{F}}$, and $\boldsymbol{b_{i}}^\intercal \in \lbrace -1,1 \rbrace^{K\times 1} $ is the hash code for the $i$th image. Eq. \ref{similaritynew} could be further simplified into the Eq. \ref{similaritynew2} ,
\begin{equation}\label{similaritynew2}
\underset{\boldsymbol{B}}{\mathrm{\textit{arg} min}} \hspace{0.1cm} tr(\boldsymbol{B}^\intercal \boldsymbol{L^{S}}\boldsymbol{B})
\end{equation} 
where $\boldsymbol{L^{S}}$ is the Laplacian matrix computed as $\boldsymbol{L^{S}} = \boldsymbol{D^{S}} - \boldsymbol{W^{S}}$. Here $\boldsymbol{D^{S}}$ is a $N \times N$ diagonal degree matrix whose entries are given by ${D^{S}_{ii}} =  \sum^{N}_{j = 1} {W^{S}_{ij}}$. 

By minimizing the energy function given in Eq. \ref{similaritynew3}, we preserve between class similarity information.
\begin{equation}\label{similaritynew3}
\sum_{(i,j)} \frac{{W^{inter}_{ij}}}{4K}||\boldsymbol{b_{i}}^\intercal - \boldsymbol{b_{j}}^\intercal ||^{2}_{2}
\end{equation} 
The Eq. \ref{similaritynew3} could be further simplified into the Eq.\ref{similaritynew4} ,
\begin{equation}\label{similaritynew4}
\underset{\boldsymbol{B}}{\mathrm{\textit{arg} min}} \hspace{0.1cm} tr(\boldsymbol{B}^\intercal \boldsymbol{L^{inter}}\boldsymbol{B})
\end{equation} 
where $\boldsymbol{L^{inter}}$ is the Laplacian matrix computed as $\boldsymbol{L^{inter}} = \boldsymbol{D^{inter}} - \boldsymbol{W^{inter}}$. Here $\boldsymbol{D^{inter}}$ is a $N \times N$ diagonal degree matrix whose entries are given by ${D^{inter}_{ii}} =  \sum^{N}_{j = 1} {W^{inter}_{ij}}$. 
The Laplacian matrix $\boldsymbol{L}$ encoding the information of intra modal similarity is given as in Eq. \ref{mainlaplace}.
\begin{equation} \label{mainlaplace}
\boldsymbol{L} = \boldsymbol{L^{S}} - \alpha \boldsymbol{L^{inter}}
\end{equation}
where $\alpha$ is a trade off parameter  for balancing the scale of $\boldsymbol{L^{S}}$ and $\boldsymbol{L^{inter}}$. Thus our aim to preserve the intra-modal similarity could be achieved by minimizing the objective function in \ref{similaritymain}:
\begin{equation} \label{similaritymain}
\underset{\boldsymbol{B}}{\mathrm{\textit{arg} min}} \hspace{0.1cm} tr(\boldsymbol{B}^\intercal \boldsymbol{L}\boldsymbol{B})
\end{equation}

\subsection{Optimization}
Combining  Eq. \ref{similaritymain} and Eq. \ref{firstobjectivefunction}, the final objective function can be formed as shown in Eq. \ref{finalobjectivefunction} 
\begin{equation}\label{finalobjectivefunction}
\begin{split}
\underset{\boldsymbol{B, W_{img}, W_{txt}}}{\mathrm{\textit{arg} min}} \hspace{0.05cm} ||\boldsymbol{Y}^\intercal \boldsymbol{W_{txt}} - sgn(\boldsymbol{X}^\intercal \boldsymbol{W_{img}})||^{2} + \beta ||\boldsymbol{W_{txt}}||^{2} + \gamma tr(\boldsymbol{B}^\intercal \boldsymbol{L} \boldsymbol{B})  \hspace{1.3cm}          
\end{split}
\end{equation}
Eq. \ref{finalobjectivefunction} could be further simplified by using the hash code matrix $\boldsymbol{B}$ instead of $sgn(\boldsymbol{X}^\intercal \boldsymbol{W_{img}})$. Rewriting Eq. \ref{finalobjectivefunction} with this substitution we get:
\begin{equation}\label{finalobjectivefunction2}
\begin{split}
\underset{\boldsymbol{B,W_{txt}}}{\mathrm{\textit{arg} min}} \hspace{0.05cm} ||\boldsymbol{Y}^\intercal \boldsymbol{W_{txt}} - \boldsymbol{B}||^{2} + \beta ||\boldsymbol{W_{txt}}||^{2} + \gamma tr(\boldsymbol{B}^\intercal \boldsymbol{L}\boldsymbol{B})  \hspace{1.3cm} \\            
\end{split}
\end{equation}
Differentiating the Eq. \ref{finalobjectivefunction2} with respect to $\boldsymbol{W_{txt}}$ and equating it to zero we obtain the following Eq.:
\begin{equation} \label{Wtxt matrix}
\begin{split}
2\boldsymbol{Y}\boldsymbol{Y}^\intercal \boldsymbol{W_{txt}} - 2\boldsymbol{Y}\boldsymbol{B} + 2\beta \boldsymbol{W_{txt}} = 0 \\
\boldsymbol{W_{txt}} = (\boldsymbol{Y}\boldsymbol{Y}^\intercal + \beta \boldsymbol{I})^{-1}\boldsymbol{YB}
\end{split}
\end{equation} 
Substituting the value of $\boldsymbol{W_{txt}}$ into the Eq. \ref{finalobjectivefunction2}, we have:
\begin{equation}
\begin{split}
\boldsymbol{Y}^\intercal \boldsymbol{W_{txt}} - \boldsymbol{B} = \boldsymbol{Y}^\intercal (\boldsymbol{Y}\boldsymbol{Y}^\intercal + \beta \boldsymbol{I})^{-1}\boldsymbol{YB} - \boldsymbol{B}\\
= \boldsymbol{Y}^\intercal \boldsymbol{MYB} - \boldsymbol{B} \hspace{1.9cm}
\end{split}
\end{equation} 
where, $\boldsymbol{M} = (\boldsymbol{Y}\boldsymbol{Y}^\intercal + \beta \boldsymbol{I})^{-1}$. Thus, the Eq. \ref{finalobjectivefunction2} could be re-written as shown in below Eq.:
\begin{equation} \label{spectralhashing}
\begin{split}
\underset{\boldsymbol{B}}{\mathrm{\textit{arg} min}} \hspace{0.05cm} tr(\boldsymbol{B}^\intercal (\boldsymbol{I} - \boldsymbol{Y}^\intercal \boldsymbol{M}\boldsymbol{Y} + \gamma \boldsymbol{L}) \boldsymbol{B})\\
s.t. \hspace{0.2cm} \boldsymbol{B} \in \lbrace -1,+1\rbrace^{(N \times K)} \hspace{1.2cm}
\end{split}
\end{equation}
Replacing $(\boldsymbol{I} - \boldsymbol{Y}^\intercal \boldsymbol{M}\boldsymbol{Y} + \gamma \boldsymbol{L})$ with the matrix $\boldsymbol{C}$, in Eq. \ref{spectralhashing}, we have:
\begin{equation} \label{spectralhashing2}
\begin{split}
\underset{\boldsymbol{B}}{\mathrm{\textit{arg} min}} \hspace{0.05cm} tr(\boldsymbol{B}^\intercal \boldsymbol{C}\boldsymbol{B}) \hspace{2.65cm}\\
s.t. \hspace{0.2cm} \boldsymbol{B} \in \lbrace -1,+1\rbrace^{(N \times K)} \hspace{1.2cm}\\
\end{split}
\end{equation}
where $\boldsymbol{I}$ is the identity matrix of size $(K \times K)$.
Derivations of the Eq. \ref{spectralhashing} and  Eq. \ref{finalobjectivefunction2} are given in the Appendix.\\
Eq. \ref{spectralhashing2} could be solved efficiently by removing the constraint $\boldsymbol{B} \in \lbrace - 1, 1 \rbrace^{(N \times K)}$. The solution of the objective function is simply the $K$ thresholded eigenvectors of $\boldsymbol{C}$ corresponding to the $K$ smallest non-zero eigenvalues \cite{weiss2009spectral}. 

\subsection{Learning the hash function}
Note that we still have not obtained the hash function $f(\boldsymbol{x_{i}}) = sgn(\boldsymbol{X}^\intercal \boldsymbol{W_{img}})$, i.e. our objective of obtaining $\boldsymbol{W_{img}}$ has still not been achieved. Previous works have tried to minimize the quantization loss $ || \boldsymbol{B} - \boldsymbol{X}^\intercal \boldsymbol{W_{img}} ||^{2}$ to obtain $\boldsymbol{W_{img}}$ (\cite{gong2011iterative}, \cite{shen2015learning}). However, in the proposed method hash codes are obtained by taking the sign of $(\boldsymbol{X}^\intercal \boldsymbol{W_{img}})$. Hence, as suggested in already published work \cite{kang2016column}, relaxing the discrete constraint and minimizing the quantization loss will not be suitable for the proposed method.
Instead, inspired by \cite{rastegari2012attribute}, we leverage support vector machines and use hinge loss, which is endowed with the max-margin property in hyperplane learning to obtain $\boldsymbol{W_{img}}$. Each column of $\boldsymbol{W_{img}}$ could be viewed as a split in the visual space. 
Therefore, we minimize the energy formulated in Eq. \ref{supportvectors} 
\begin{equation} \label{supportvectors}
\sum^{K}_{k = 1} ||\boldsymbol{w_{img}^{k}}||^{2} + \lambda \sum^{K}_{k = 1}\sum^{N}_{i = 1} \max(0, 1-b^{k}_{i}(\boldsymbol{x_{i}}^\intercal \boldsymbol{w_{img}^{k}}))
\end{equation}
where $\boldsymbol{w_{img}^{k}} \in \mathbb{R}^{(d_{x} \times 1)}, k \in \lbrace 1,2, \ldots K \rbrace$ is the $k$th column of the projection matrix $\boldsymbol{W_{img}}$ in Eq. \ref{firstobjectivefunction}. The hyper parameter $\lambda$ balances the hyperplane margin and the empirical training error. The proposed algorithm is summarized in the \textbf{Algorithm 1}. We used the LIBLINEAR library \cite{fan2008liblinear} for training the max-margin classifier.

\begin{algorithm} \label{algorithm1}
\textbf{Input:} Training images $\boldsymbol{X} = [\boldsymbol{x_{1}} , \boldsymbol{x_{2}},..., \boldsymbol{x_{N}}] \in \mathbb{R}^{(d_{x} \times N)}$, textual information matrix $\boldsymbol{Y} = [\boldsymbol{y_{1}} , \boldsymbol{y_{2}},..., \boldsymbol{y_{N}}] \in \mathbb{R}^{(d_{y} \times  N)}$ and class labels for each of the training samples in $\boldsymbol{X}$.\\
\textbf{Output:} $\boldsymbol{B} \in \lbrace -1, +1 \rbrace^{N\times K} \hspace{0.2cm}, \boldsymbol{W_{img}}, \hspace{0.2cm} \boldsymbol{W_{txt}} $\\
\textbf{Steps Involved:}\\
1. Formulate the objective function in Eq. \ref{finalobjectivefunction} in the form of Eq. \ref{spectralhashing2}.\\
2. Solve the Eq. \ref{spectralhashing2} by exploiting the smallest K eigenvectors to obtain the hash code matrix $\boldsymbol{B}$.\\
3. Obtain the textual projection matrix $\boldsymbol{W_{txt}}$ using the Eq. \ref{Wtxt matrix}.\\
4. Obtain $\boldsymbol{W_{img}}$ using the hash code matrix $\boldsymbol{B}$.\\
5. Continue iterations to obtain $\boldsymbol{W_{img}}$ from $\boldsymbol{B}$ until change in $\boldsymbol{W_{img}}$ in successive iterations is negligible.\\ 
6. Repeat steps 4 and 5 iteratively until the convergence.\\
\textbf{End} 
\caption{Training Algorithm}
\end{algorithm}

\subsection{Domain Adaptation}

Though zero shot learning as a whole is not a domain adaptation problem but its main objective is to learn a projection function for target domain onto the same semantic space using the concepts learnt from the source data. This problem of projection domain shift in the context of zero shot learning was first studied in \cite{fu2014transductive}.
Traditional zero shot learning approaches, the learned cross-modal embedding matrix $\boldsymbol{W_{img}}$ from the seen classes is applied directly to the unseen classes. Since the training classes and testing ones are disjoint, the learned parameters may not be seamlessly suitable for unseen classes. The problem of domain shift was introduced again in \cite{kodirov2015unsupervised}, in which authors introduced dictionary learning methods to address the issue. In our work, in order to produce the correct hash codes for images from seen and unseen catagories, we further propose a new domain adaptation strategy to learn an improved transformation matrix $\boldsymbol{W^{*}_{img}}$ for the unseen classes. Domain adaptation technique has proven its effectiveness in applications where there is a lot of training data in one domain but little to none in another (\cite{margolis2011literature}, \cite{kodirov2015unsupervised}, \cite{gan2016learning}). We formulate the problem of domain adaptation in online learning fashion. Practically, to ensure a more reliable gradient estimate, instead of using a single sample at a time to update $\boldsymbol{W_{img}^{*}}$, we use a (mini) batch gradient descent to consider more samples while maintaining the efficiency at each iteration. In our experiments, the mini-batch is selected as 20 samples and the update usually takes 3-4 iterations.\\
Given feature matrix $\boldsymbol{X_{test}} \in \mathbb{R}^{(d_{x} \times N_{test})}$ representing a mini-batch of images belonging to one of the unseen classes, we first project it to the class embedding space with the transformation matrix $\boldsymbol{W_{img}}$ learned from the seen classes, such that its label can be predicted by a nearest neighbour classifier with inner-product similarity. We also want to obtain the hash code matrix $\boldsymbol{B^{*}}$ for the mini batch of images. We then obtain the new improved transformation matrix $\boldsymbol{W^{*}_{img}}$ by optimizing the Eq. \ref{finaldomainadpatation}:
\begin{equation}\label{finaldomainadpatation}
\begin{split}
\underset{\boldsymbol{B^{*}, W_{img}^{*}}}{\mathrm{\textit{arg} min}}  \sum_{i = 1}^{N_{test}}\sum_{j=1}^{n_{u}} U_{ij}||\boldsymbol{p_{j}} - \boldsymbol{b_{i}^{*}}||^{2} + \beta_{1} ||\boldsymbol{W_{img}^{*}} - \boldsymbol{W_{img}}||^{2} \\
 + \lambda_{1} \sum^{K}_{k=1}\sum^{N_{test}}_{i=1} \max \Big( 0, (1 - {b_{i}^{*k}}((\boldsymbol{W_{img}^{*k}})^\intercal \boldsymbol{x_{i}}) ) \Big)
\end{split}
\end{equation}
Here, ${N_{test}}$ is the number of samples in the mini batch, $n_{u}$ is the number of unseen classes, $\boldsymbol{p_{j}}$ denotes the class prototype and is equal to $(\boldsymbol{y_{j}}^\intercal \boldsymbol{W_{txt}})$ for the class $j$, $\boldsymbol{y_{j}}$ is the textual/attribute vector corresponding to the $j$th class, $\boldsymbol{W_{img}^{*k}}$ is the $k$th column of the updated image projection matrix $\boldsymbol{W_{img}^{*}}$ and $b_{i}^{*k}$ is the $k$th bit of the updated hash code of the $i$th image of the mini batch. Our goal is to obtain the updated matrix $\boldsymbol{W_{img}^{*}}$ and the hash code matrix $\boldsymbol{B^{*}}$ for the mini batch under consideration. The first term in Eq. \ref{finaldomainadpatation} is a prototype term, which ensures that the hash code of each testing sample $\boldsymbol{x_{i}}$ is close to the corresponding class prototype (class to which it is likely to belong). The value $U_{ij}$ denotes the weight to enforce closeness between the hash code of $\boldsymbol{x_{i}}$ and the $j$th class prototype $\boldsymbol{y_{j}}$ from the set of testing (unseen) classes, i.e., $U_{ij}$ = $\phi(sgn(\boldsymbol{x_{i}}^\intercal \boldsymbol{W_{img}^{*}}),(\boldsymbol{y_{j}}^\intercal \boldsymbol{W_{txt}}))$, where $\phi(\boldsymbol{p},\boldsymbol{q})$ calculates the cosine distance between two vectors $\boldsymbol{p}$ and $\boldsymbol{q}$. Also, $\boldsymbol{U} \in \mathbb{R}^{(N_{test} \times n_{u})}$. The second term $|| \boldsymbol{W_{img}^{*}} - \boldsymbol{W_{img}} ||_{2}^{2}$ constrains the learned $\boldsymbol{W^{*}_{img}}$ for unseen classes to be similar to $\boldsymbol{W_{img}}$ learned from seen classes. Since $\boldsymbol{W_{img}}$ is learned by preserving the semantic consistency across different modalities, this transformation regularization term ensures that the learned $\boldsymbol{W^{*}_{img}}$ can also effectively project the visual feature to the class embedding space. The final term of the objective function ensures that the new projection matrix $\boldsymbol{W_{img}^{*}}$ produces hash codes in the most efficient way.

\subsubsection{Optimization with respect to $\boldsymbol{B^{*}}$}
We need to optimize the following objective function with respect to $\boldsymbol{B^{*}}$.
\begin{equation}
\underset{\boldsymbol{B^{*}}}{\mathrm{\textit{arg} min}}  \sum_{i = 1}^{N_{test}}\sum_{j=1}^{n_{u}} U_{ij}||\boldsymbol{p_{j}} - \boldsymbol{b_{i}}^{*}||^{2} + \lambda_{1} \sum^{K}_{k=1}\sum^{N_{test}}_{i=1} \max \Big( 0, (1 - b_{i}^{*k}((\boldsymbol{W_{img}^{*k}})^\intercal \boldsymbol{x_{i}}) ) \Big)
\end{equation}
Differentiating above equation with respect to $\boldsymbol{B^{*}}$ partially, we get
\begin{equation} \label{Bderivative}
-2\boldsymbol{UP} + 2\boldsymbol{D^{u}}\boldsymbol{B^{*}} - 2\lambda_{1}\boldsymbol{F} 
\end{equation}
where, $\boldsymbol{D^{u}}$ is a $N_{test}\times N_{test}$ diagonal matrix, whose diagonal elements are given by $D^{u}_{ii} = \sum_{j}U_{ij}$. $\boldsymbol{P} \in \mathbb{R}^{(n_{u}\times K)}$ is the projection of textual/attribute information of each of the unseen classes using the textual projection matrix $\boldsymbol{W_{txt}}$  and is equal to $\boldsymbol{Y_{nu}}^\intercal \boldsymbol{W_{txt}}$. $\boldsymbol{Y_{nu}} \in \mathbb{R}^{(d_{y} \times n_{u})}$ is the matrix containing the general textual/attribute information of each of the unseen classes. Each element of $\boldsymbol{F}$ (represented as $F_{ik}$) is given in Eq. \ref{BBB} :
\begin{equation} \label{BBB}
  F_{ik} =
  \begin{cases}
    -(\boldsymbol{W_{img}^{*k}})^\intercal \boldsymbol{x_{i}} & \text{if} \hspace{0.1cm}(b_{i}^{*k}(\boldsymbol{W_{img}}^{*k})^\intercal \boldsymbol{x_{i}})< 1 \\
    0 & \text{otherwise}
  \end{cases}
\end{equation}
Equating the Eq. \ref{Bderivative} to zero, followed by thresholding we have:
\begin{equation} \label{Boptimization}
\boldsymbol{B^{*}} = sgn((\boldsymbol{D^{u}})^{-1}(\boldsymbol{UP} + \lambda_{1}\boldsymbol{F}))
\end{equation}

\subsubsection{Optimization with respect to $\boldsymbol{W_{img}^{*}}$}
To perform domain adaptation, we need to optimize the following objective function with respect to $\boldsymbol{W_{img}^{*}}$.
\begin{equation}
\underset{\boldsymbol{W_{img}^{*}}}{\mathrm{\textit{arg} min}} \hspace{0.2cm} \beta_{1} ||\boldsymbol{W_{img}^{*}} - \boldsymbol{W_{img}}||^{2} + \frac{\lambda_{1}}{N_{test}} \sum^{K}_{k=1}\sum^{Ntest}_{i=1} \max \Big( 0, (1 - b_{i}^{*k}((\boldsymbol{W_{img}^{*k}})^\intercal \boldsymbol{x_{i}}) ) \Big)
\end{equation}
Differentiating above equation with respect to $\boldsymbol{W_{img}^{*}}$, we get
\begin{equation} \label{Woptimization}
2\beta_{1} (\boldsymbol{W_{img}^{*}} - \boldsymbol{W_{img}}) + \frac{\lambda_{1}}{N_{test}} \boldsymbol{T}
\end{equation}
where, each $k$-th column of $\boldsymbol{T}$ ( represented as $\boldsymbol{T^{k}}$ ) is given as:
\begin{equation}
\boldsymbol{T^{k}} = \sum^{N_{test}}_{i = 1} G^{k}_{i}
\end{equation} 
where, 
\begin{equation}
G_{i}^{k} =
  \begin{cases}
    (-b_{i}^{*k}\boldsymbol{x_{i}}) & \text{if} \hspace{0.1cm}(b_{i}^{*k}(\boldsymbol{W_{img}^{*k}})^\intercal \boldsymbol{x_{i}})< 1 \\
    0 & \text{otherwise}
  \end{cases}
\end{equation}
Equating the above Eq \ref{Woptimization} to zero, we have:
\begin{equation} \label{Wstaroptimization2}
\boldsymbol{W_{img}^{*}} = \boldsymbol{W_{img}} - \frac{\lambda_{1}}{(2\beta_{1}N_{test})}\boldsymbol{T}
\end{equation}
Our method for online domain adaptation is summarized in \textbf{Algorithm 2}.
\begin{algorithm} \label{algorithm3}
\textbf{Input:} Mini-batch of Testing images $\boldsymbol{X} \in \mathbb{R}^{(d_{x} \times N_{test})}$, textual information matrix for each of the unseen classes $\boldsymbol{Y_{nu}} = [\boldsymbol{y_{1}} , \boldsymbol{y_{2}},..., \boldsymbol{y_{n_{u}}}] \in \mathbb{R}^{ {(d_{y}} \times n_{u})}$, learned projection matrices from seen classes $\boldsymbol{W_{img}}$ and $\boldsymbol{W_{txt}}$. \\
\textbf{Output:} $\boldsymbol{B^{*}} \in \lbrace -1, +1 \rbrace^{1\times K} \hspace{0.2cm}, \boldsymbol{W_{img}^{*}}$\\
\textbf{Steps Involved:}\\
1. $\boldsymbol{B^{*}}$: $\boldsymbol{B^{*}}$ $\leftarrow$ $\boldsymbol{B}$ $= sgn(\boldsymbol{X}^\intercal \boldsymbol{W_{img}^{*}})$.  \\ 
2. $\boldsymbol{W_{img}^{*}}$ $\leftarrow$ $\boldsymbol{W_{img}}$\\
3. Calculate the weight matrix $\boldsymbol{U} \in \mathbb{R}^{(N_{test}\times n_{u})}$.\\
4. Fix $\boldsymbol{W_{img}}^{*}$ to optimize $\boldsymbol{B^{*}}$ with Eq \ref{Boptimization}\\
5. Fix $\boldsymbol{B^{*}}$ to optimize $\boldsymbol{W_{img}^{*}}$ with Eq \ref{Wstaroptimization2}\\
6. Update $\boldsymbol{B^{*}}$: $\boldsymbol{B^{*}}$ $\leftarrow$ $sgn(\boldsymbol{X}^\intercal \boldsymbol{W_{img}^{*}})$\\
7. Repeat steps 3 to 6 iteratively until the convergence.\\
\textbf{End} 
\caption{Optimization for Domain Adaptation}
\end{algorithm}

\section{Experiments}
\subsection{Datasets}
In our experiments, we employ three real-life image datasets, CIFAR-10 dataset \cite{krizhevsky2009learning}, CUB-200-2011 Birds (CUB) dataset \cite{WahCUB_200_2011} and Animals with Attributes dataset (AwA) \cite{lampert2009learning}. First, we present the experimental settings, then describe the datasets and finally report the experimental results.

\subsection{Experimental Settings}
To quantitatively evaluate our algorithm, we used two metrics i.e., Mean Average Precision (MAP) \cite{turpin2006user} and the Precision-Recall curve calculated among the range of whole database as measurements. 
MAP focuses on the ranking of retrieval results. We compared the proposed algorithm with four state-of-the-art hashing approaches, three of them being supervised algorithms: COSDISH \cite{kang2016column}, SDH\cite{shen2015supervised} and ZSH \cite{yang2016zero} and one unsupervised hashing algorithm: IMH \cite{shen2013inductive}. To evaluate the performance and efficiency of our domain adaptation algorithm (OURS-DA), we also present the results of the hashing algorithm without using domain adaptation (OURS). In all experiments, irrespective of dataset, 2,500 images from the unseen category were selected as query images, and the remaining test images together with the images of seen categories were combined to form the retrieval database. We illustrate the performance our our method with all the comparing approaches with respect to hash code length of size 8, 16, 32, 64, 96, 128, 192, 256 and 512.

\textbf{Parameter Settings:} For IMH and ZSH algorithm, we randomly sampled 1,000 anchors from the training dataset. For all comparing approaches, we followed the parameter settings as suggested by the authors of the corresponding papers. We tuned our experimental parameters using grid-search with cross validation. We empirically set the regularization parameters $\alpha$ to 0.2, $\beta$ to 6 $\times$ $10^{-6}$, $\gamma$ to $10$, $\lambda$ to 0.2, $\lambda_{1}$ to 0.1, $\beta_{1}$ to 0.1, $\sigma$ to 0.5 and set the number of $k$ nearest neighbours to 10.    

\subsection{Datasets}

\begin{figure}
\centering 
\subfigure{\includegraphics[width= 0.5\textwidth]{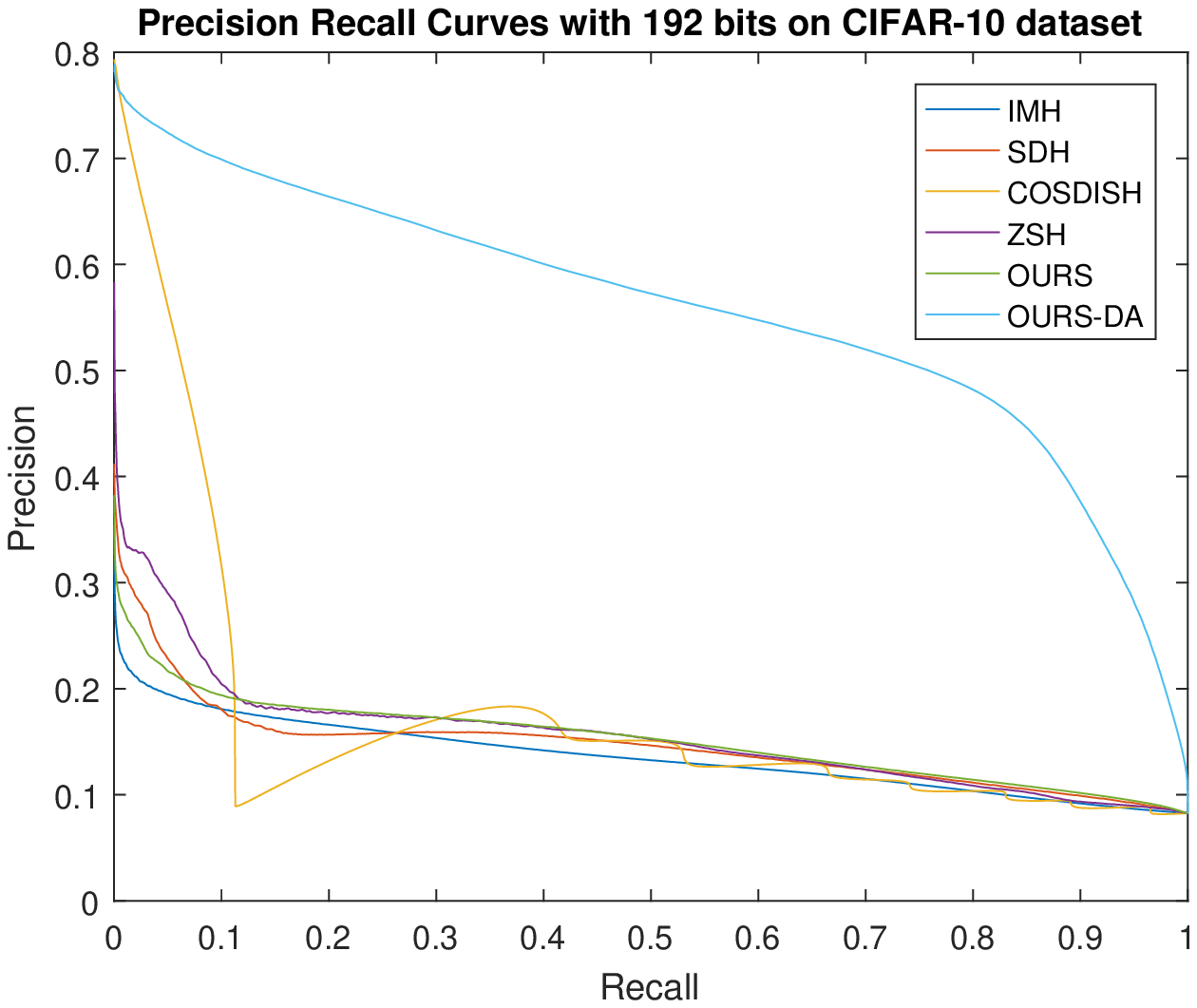}}%
\subfigure{\includegraphics[width= 0.5\textwidth]{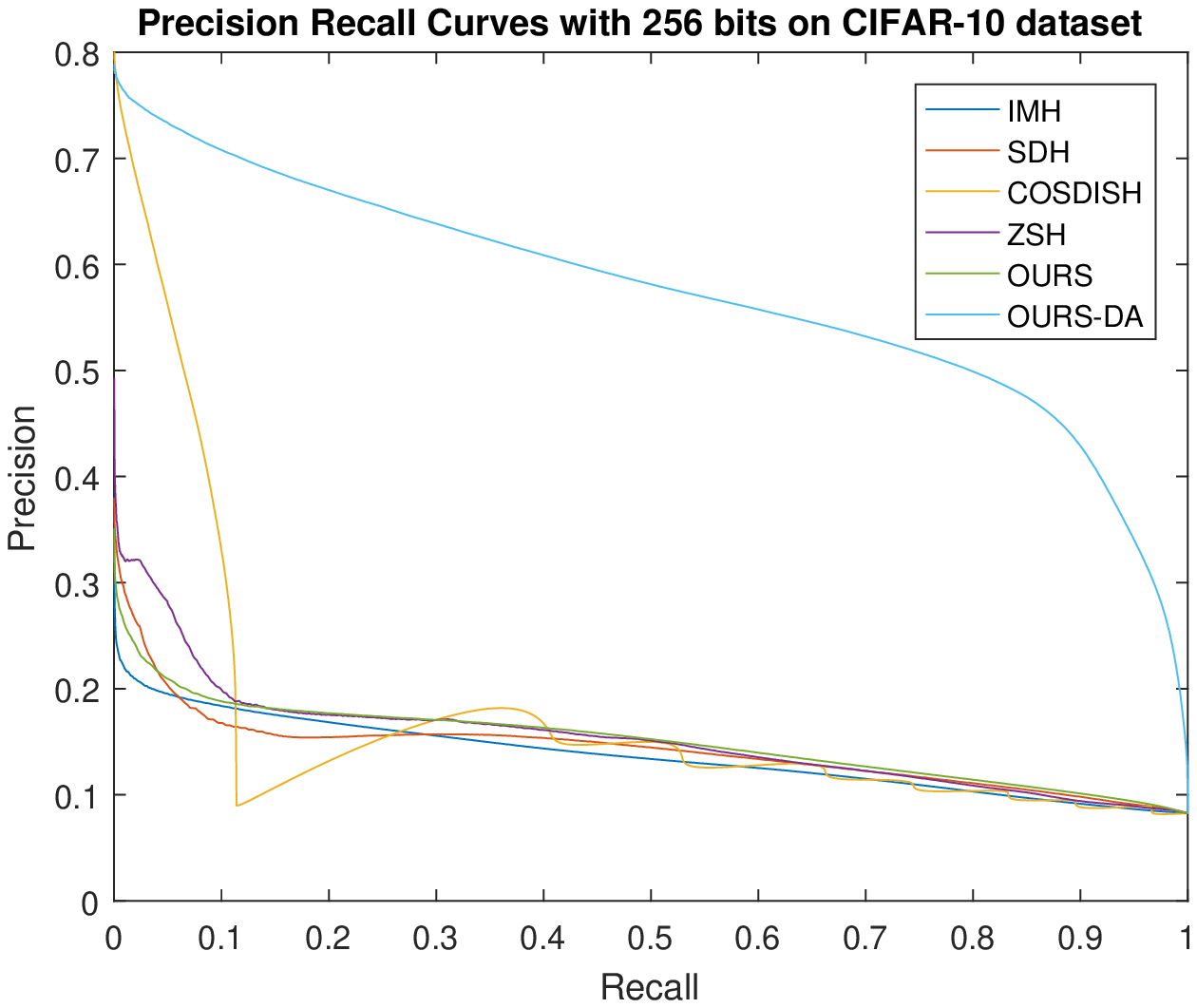}}%
\caption{Comparison with state-of-the-art binary code learning methods for image retrieval with two different code lengths in Precision-Recall curve on CIFAR-10 dataset. \label{precisionrecallCIFAR}}
\end{figure}

\textbf{CIFAR-10 Dataset:} CIFAR-10 dataset consists of 60,000 images which are divided into 10 classes with 6,000 samples in each class. The classes are mutually exclusive. We used gist features \cite{oliva2001modeling} for CIFAR-10 dataset to form a feature vector of dimension 512 for each image. We randomly split the dataset into images of 8 categories of objects as seen classes and the images from the rest of the 2 classes form the unseen classes. Corresponding to each class in the dataset, we extracted the 50-dimensional semantic vectors from the Huang dataset \cite{huang2012improving}.
We show the comparison results in MAP and Precision-Recall curve with different code lengths for CIFAR-10 dataset in Table. \ref{TableCIFAR} and Fig. \ref{precisionrecallCIFAR} respectively.

\begin{table}[h]
\caption{Comparison of our hashing algorithm with other hash learning methods with different code lengths in MAP on CIFAR-10. \label{TableCIFAR}}
\resizebox{\textwidth}{!}{%
\centering 
\begin{tabular}{ |c|c|c|c|c|c|c|c|c|c|  }
 \hline
 Method & 8 bits & 16 bits & 32 bits & 64 bits & 96 bits & 128bits & 192 bits & 256 bits & 512 bits\\
 \hline
IMH & 0.1848  & 0.1975 & 0.2035 & 0.2087  & 0.2172 & 0.2233 & 0.2320  & 0.2169 & 0.2108\\
 SDH & 0.1660  & 0.1715 & 0.2022 & 0.2136  & 0.2214 & 0.2396 & 0.2796  & 0.2603 & 0.2552 \\ 
 COSDISH & 0.1677 & 0.1721  & 0.2036 & 0.2164 & 0.2270  & 0.2424 & 0.2819 & 0.2736 & 0.2607 \\ 
 ZSH & 0.1769  & 0.1921 & 0.2168 & 0.2223  & 0.2430 & 0.2663 & 0.3199  & 0.2586 & 0.2358\\
 \hline
 Ours& 0.2634  & 0.2290 & 0.2018 & 0.1954  & 0.1927 & 0.1875 & 0.1872  & 0.1853 & 0.1880\\
 Ours-DA & $\boldsymbol{0.4130}$  & $\boldsymbol{0.4126}$ & $\boldsymbol{0.4313}$ & $\boldsymbol{0.4580}$  & $\boldsymbol{0.5325}$ & $\boldsymbol{0.5529}$ & $\boldsymbol{0.5881}$  & $\boldsymbol{0.6039}$ & $\boldsymbol{0.6216}$ \\
 \hline
\end{tabular}}
\end{table}

\begin{figure}
\centering
\subfigure{\includegraphics[width= 0.5\textwidth]{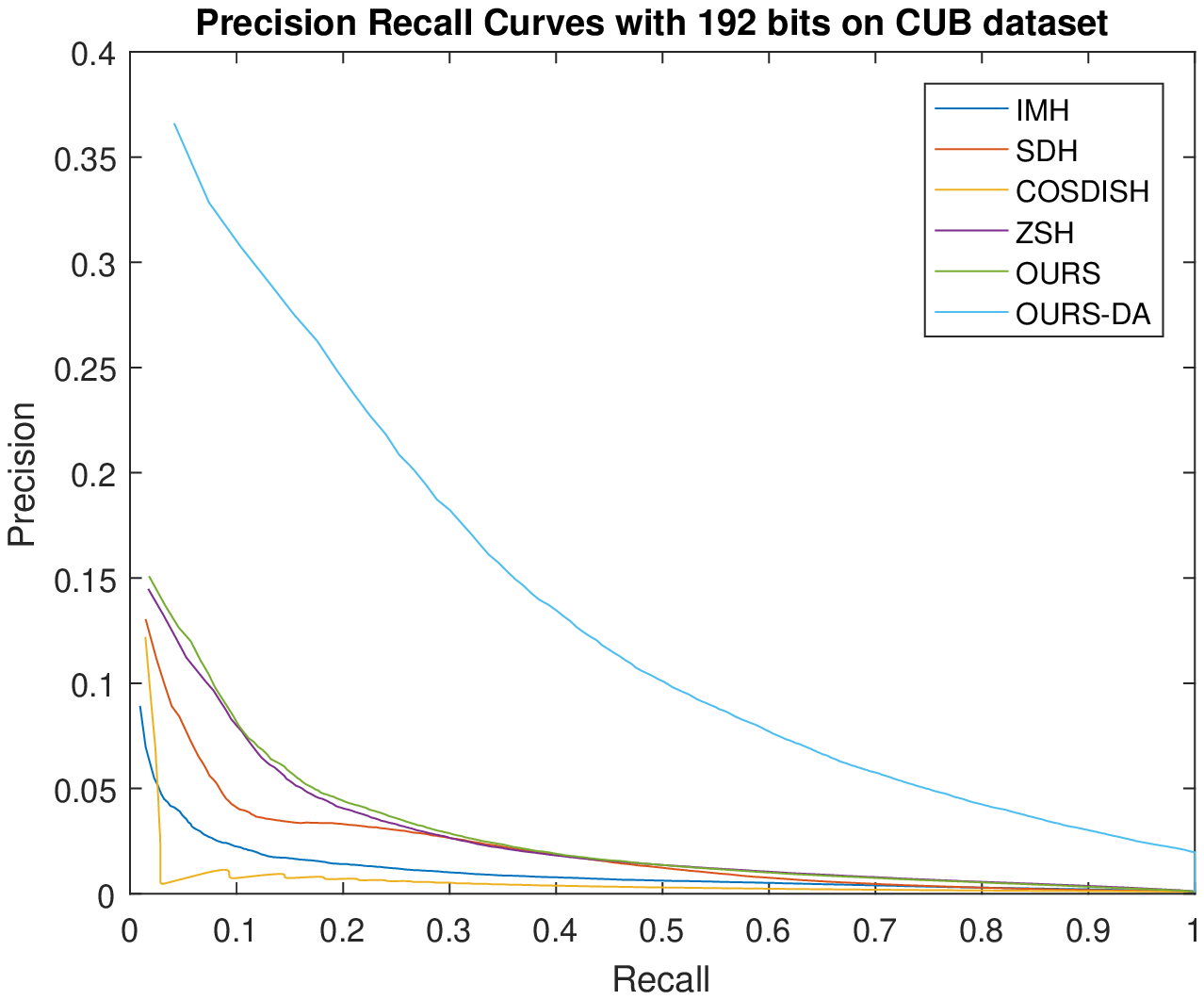}}%
\subfigure{\includegraphics[width= 0.5\textwidth]{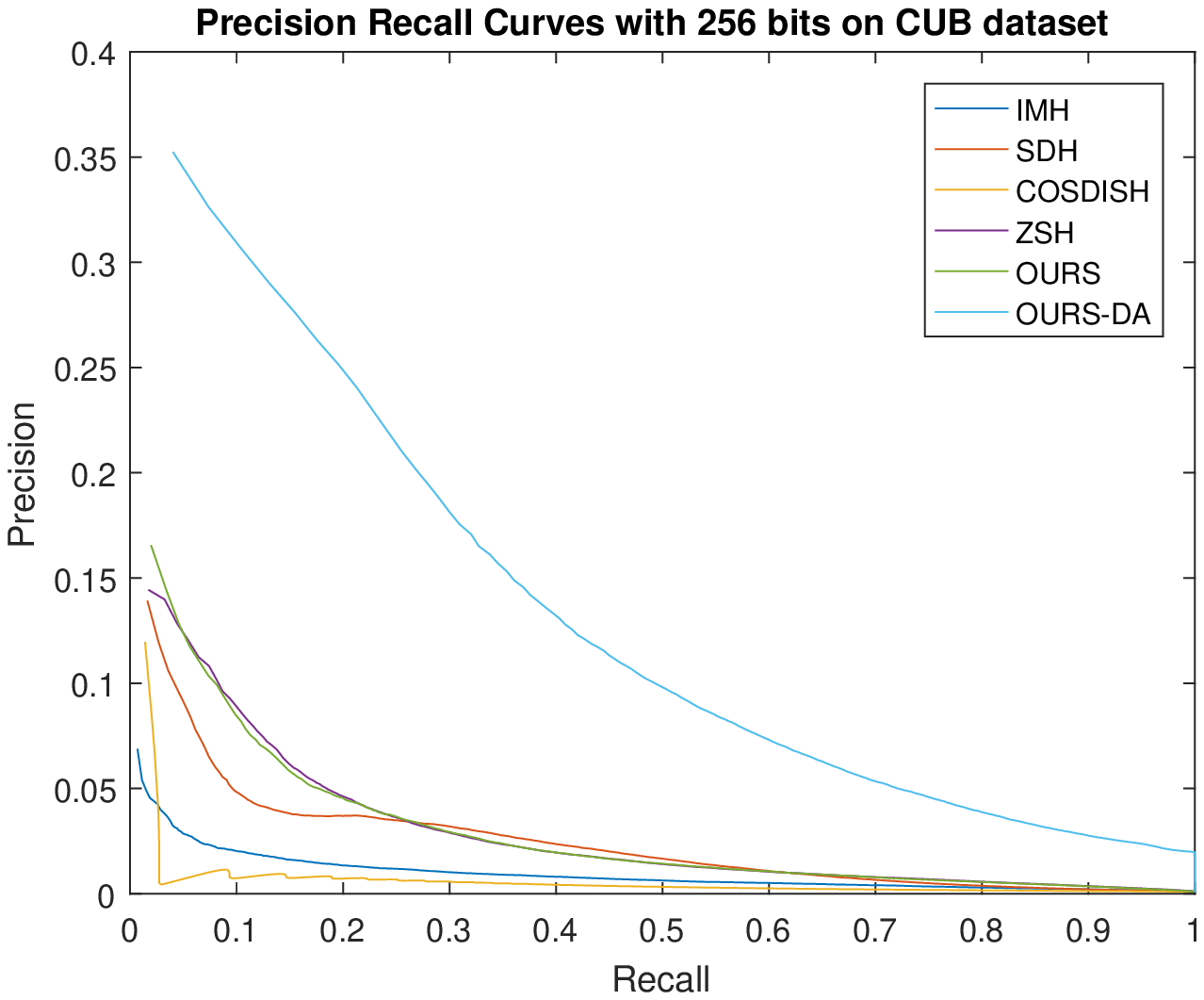}}%
\caption{Comparison with state-of-the-art binary code learning methods for image retrieval with two different code lengths in Precision-Recall curve on CUB dataset.\label{PrecisionrecallCUB}}
\end{figure}

\textbf{CUB-200-2011 Dataset:} CUB-200-2011 contains 11,788 images of 200 categories of bird subspecies with 312 fine-grained attributes such as color/shape/texture of body parts. The attribute features were preprocessed to have zero mean. For CUB dataset, we obtained deep features using VGG with 19-layer network \cite{simonyan2014very} using MatConvNet \cite{vedaldi2015matconvnet}. The image features were normalized so that each datapoint would have a unit norm. For comparing our algorithm with state of the art hashing algorithms, we followed \cite{akata2013label} to use 150 birds species as seen classes for training and the rest 50 species as unseen classes for testing. Corresponding to each of the class category, we used the preprocessed 312 fine grained attributes as our semantic vector representing the class. The comparison results are shown in MAP and Precision-Recall curve with different code lengths for CUB-200-2011 dataset in Table. \ref{TableCUB} and Fig. \ref{PrecisionrecallCUB} respectively.

\begin{table}[h]
\caption{Comparison of our hashing algorithm with other hash learning methods with different code lengths in MAP on CUB Dataset. \label{TableCUB}}

\resizebox{\textwidth}{!}{%
\centering
\begin{tabular}{ |c|c|c|c|c|c|c|c|c|c|  }
 \hline
 Method & 8 bits & 16 bits & 32 bits & 64 bits & 96 bits & 128 bits & 192 bits & 256 bits & 512 bits\\
 \hline
  IMH & 0.0330  & 0.0331 & 0.0343 & 0.0362 & 0.0435  & 0.0463   & 0.0465 & 0.0458 & 0.0449\\
 SDH & 0.0336  & 0.0383 & 0.0583 & 0.0673  & 0.0712 & 0.0585 & 0.0666  & 0.0751 & 0.0831\\
 COSDISH & 0.0074  & 0.0087 & 0.0141 & 0.0212  & 0.0267 & 0.0289 & 0.0286  & 0.0287 & 0.0266\\
 ZSH & $\boldsymbol{0.0857}$  & $\boldsymbol{0.0660}$ & 0.0581 & 0.0700  & 0.0802 & 0.0736 & 0.0855  & 0.0898 & 0.0923\\
 \hline
 Ours& 0.0386  & 0.0290 & 0.0401 & 0.0591  & 0.0705 & 0.0755 & 0.0875  & 0.0903 & 0.1021\\
 Ours-DA & 0.0383  & 0.0587 & $\boldsymbol{0.0844}$ & $\boldsymbol{0.1051}$  & $\boldsymbol{0.1135}$ & $\boldsymbol{0.1570}$ & $\boldsymbol{0.2306}$  & $\boldsymbol{0.2290}$ & $\boldsymbol{0.2049}$ \\
 \hline
\end{tabular}}
\end{table}

\begin{figure}
\centering
\subfigure{\includegraphics[width= 0.5\textwidth]{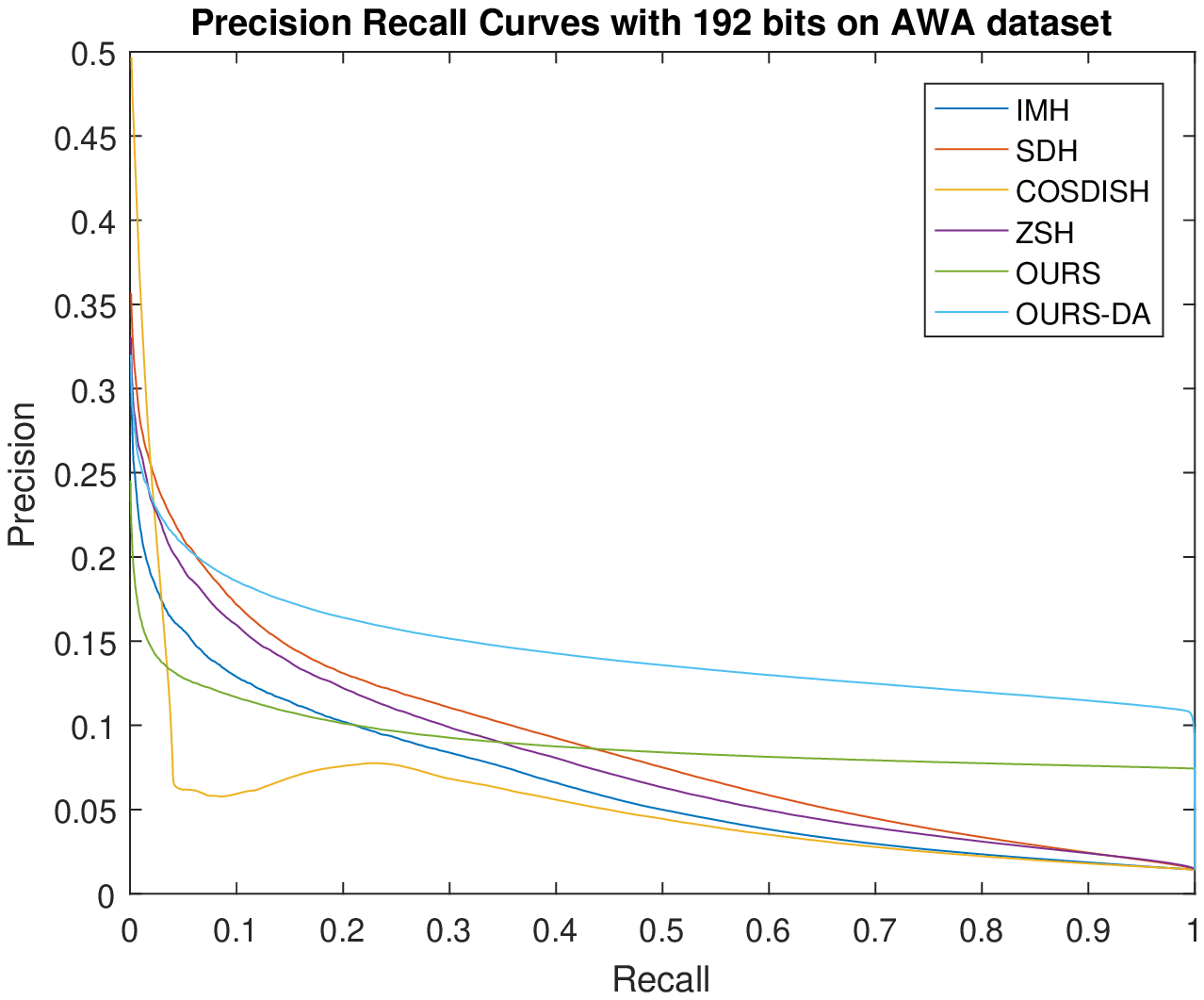}}%
\subfigure{\includegraphics[width= 0.5\textwidth]{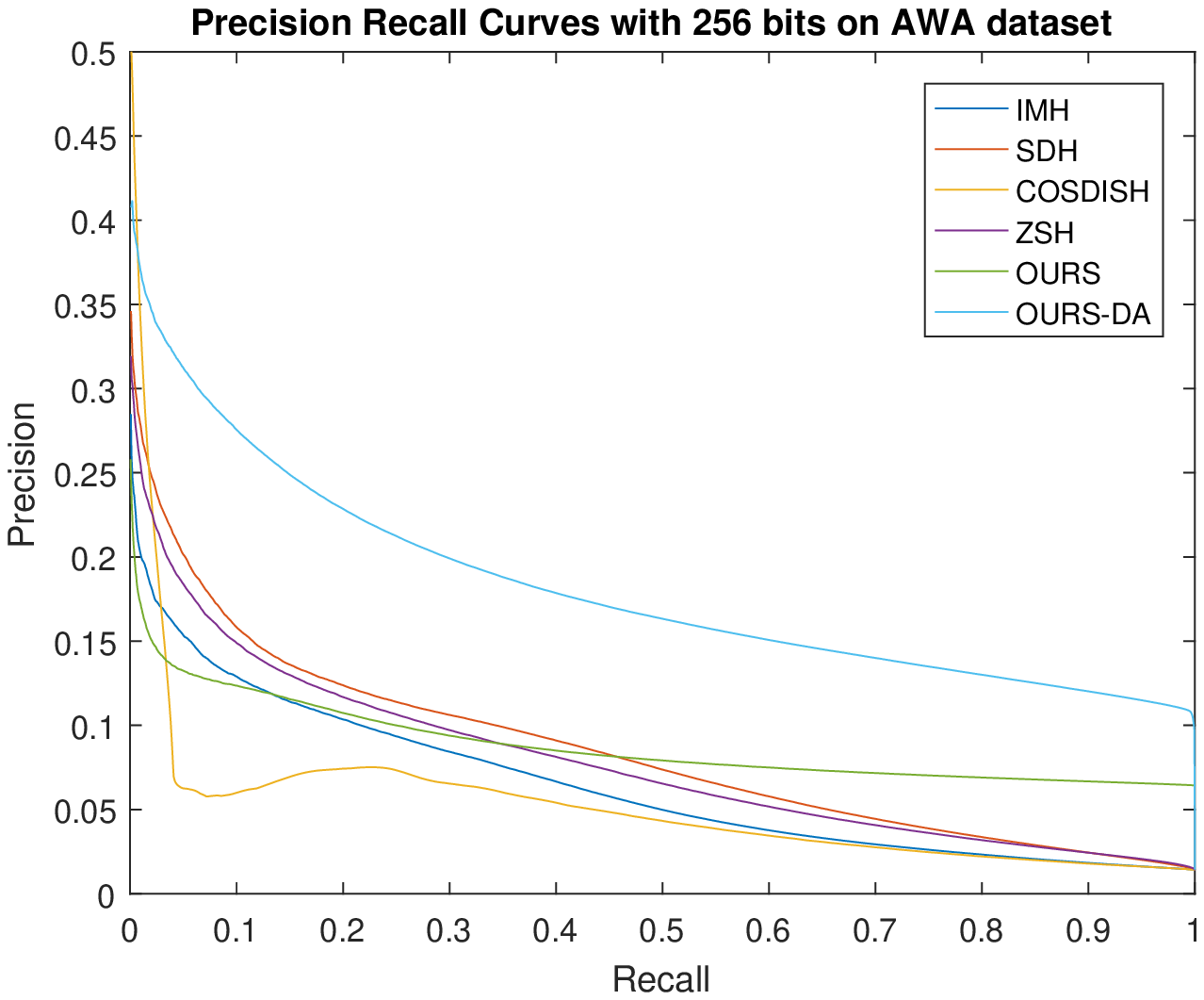}}%
\caption{Comparison with state-of-the-art binary code learning methods for image retrieval with two different code lengths in Precision-Recall curve on AwA dataset.\label{precisionrecallAWA}}
\end{figure}

\textbf{AwA Dataset:} AwA consists of 30,475 images of 50 mammals classes with 85 binary attributes including color, skin texture, body size, body part, affordance, food source, habitat, and behaviour. We used DeCAF \cite{donahue2014decaf} features for AwA dataset. We randomly split the dataset into images of 40 categories of mammals as seen classes and the images from the rest of the 10 classes form the unseen classes.  Corresponding to each of the class category, we used the 85 binary attributes as our semantic vector representing the class. We show the comparison results in MAP and Precision-Recall curve with different code lengths for CUB-200-2011 dataset in Table. \ref{TableAWA} and Fig. \ref{precisionrecallAWA} respectively.

\begin{table}[h]
\caption{Comparison of our hashing algorithm with other hash learning methods with different code lengths in MAP on AwA Dataset. \label{TableAWA}}
\resizebox{\textwidth}{!}{%
\centering
\begin{tabular}{ |c|c|c|c|c|c|c|c|c|c|  }
 \hline
 Method & 8 bits & 16 bits & 32 bits & 64 bits & 96 bits & 128bits & 192 bits & 256 bits & 512 bits\\
 \hline
 IMH & 0.1077 & 0.1082 & 0.1092 & 0.1102  & 0.1102  & 0.1103  & 0.1115  & 0.1004 & 0.0969\\
 SDH & 0.1097  & 0.0948 & 0.0743 & 0.1127  & 0.1266 & 0.1135 & 0.1277  & 0.1218 & 0.1233\\
 COSDISH & 0.0762  & 0.0779 & 0.1045 & 0.1184  & 0.1115 & 0.1035 & 0.0901  & 0.0895 & 0.0799\\
 ZSH & $\boldsymbol{0.1196}$  & $\boldsymbol{0.1087}$ & 0.1051 & 0.1219  & 0.1171 & 0.1208 & 0.1117  & 0.1069 & 0.1279\\
 \hline
 Ours& 0.0654  & 0.0834 & 0.1067 & 0.1148  & 0.1278 & 0.1281 & 0.1334  & 0.1487 & 0.1564\\
 Ours-DA & 0.0846  & 0.1030 & $\boldsymbol{0.1261}$ & $\boldsymbol{0.1375}$  & $\boldsymbol{0.1432}$ & $\boldsymbol{0.1432}$ & $\boldsymbol{0.1579}$  & $\boldsymbol{0.2063}$ & $\boldsymbol{0.1856}$ \\
 \hline
\end{tabular}}
\end{table}

\textbf{Discussion:} Tables (\ref{TableCIFAR}, \ref{TableCUB}, \ref{TableAWA}) and Figures (\ref{precisionrecallCIFAR}, \ref{PrecisionrecallCUB}, \ref{precisionrecallAWA}) prove the superiority of our algorithm over other competitive algorithms and verifies the effectiveness of the unsupervised domain adaptation in the hashing framework. Interesting observation is that IMH (unsupervised hashing algorithm) performs better or comparative to some of the supervised hashing algorithms, in terms of MAP performance, for the case of small bit size. One possible reason for this is that IMH encodes images solely with the distributional properties in the feature space whereas supervised methods exploits a transformation matrix to embed the information of visual features distribution into hash codes using independent semantic labels in the learning process. Therefore, for supervised hashing algorithms more code length is required to ensure the discriminative and descriptive power. Moreover, we could also observe that with increase in hash code length, MAP performance decreases especially in the case of using 512 bits as hash code length instead of 256 bits. This is because Hamming spaces become sparser as we increase the hash code length. 

\subsection{Effect of Seen Category Ratio on CIFAR-10}

\begin{figure}
\centering
\subfigure{\includegraphics[width= 0.5\textwidth]{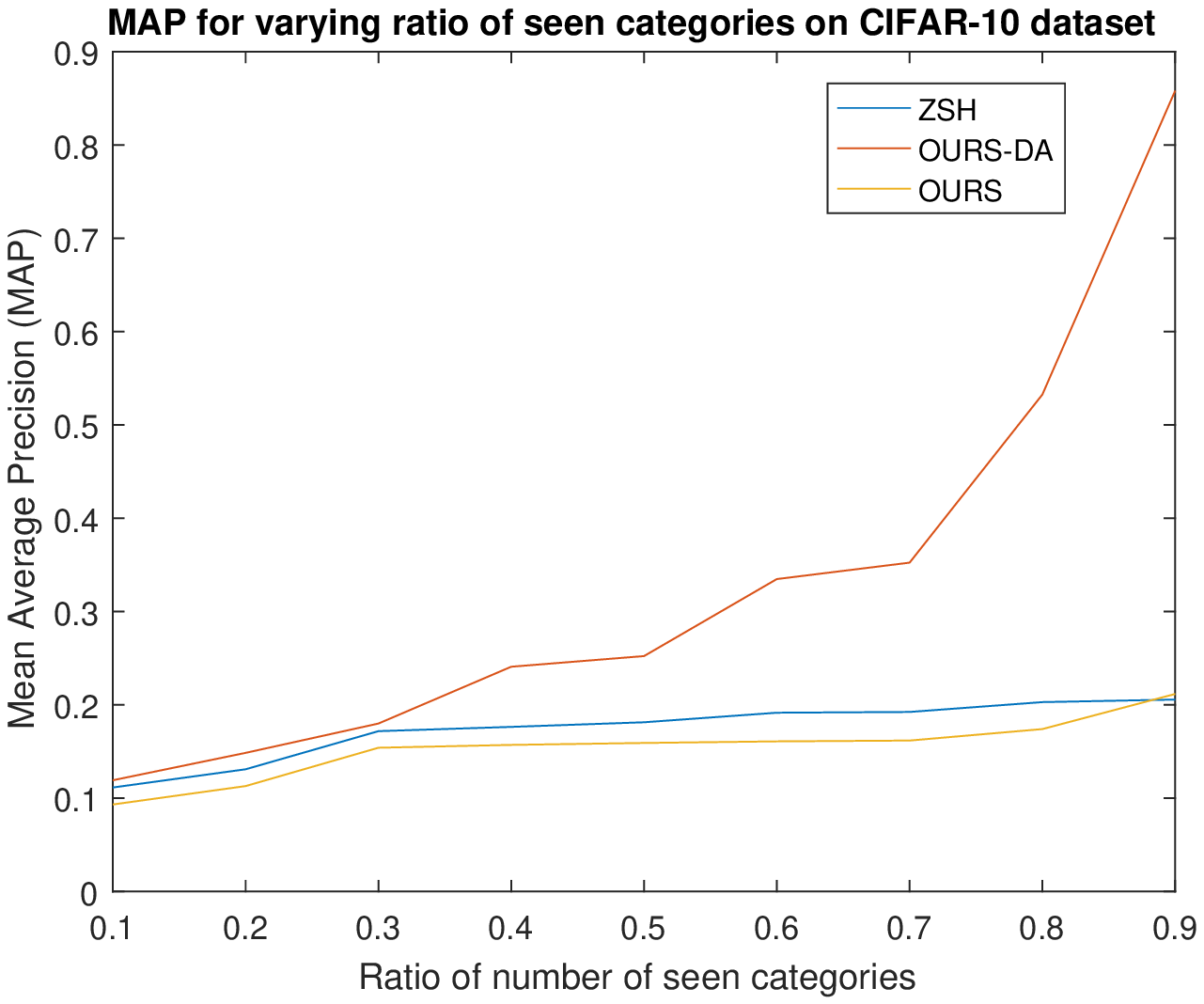}}%
\subfigure{\includegraphics[width= 0.5\textwidth]{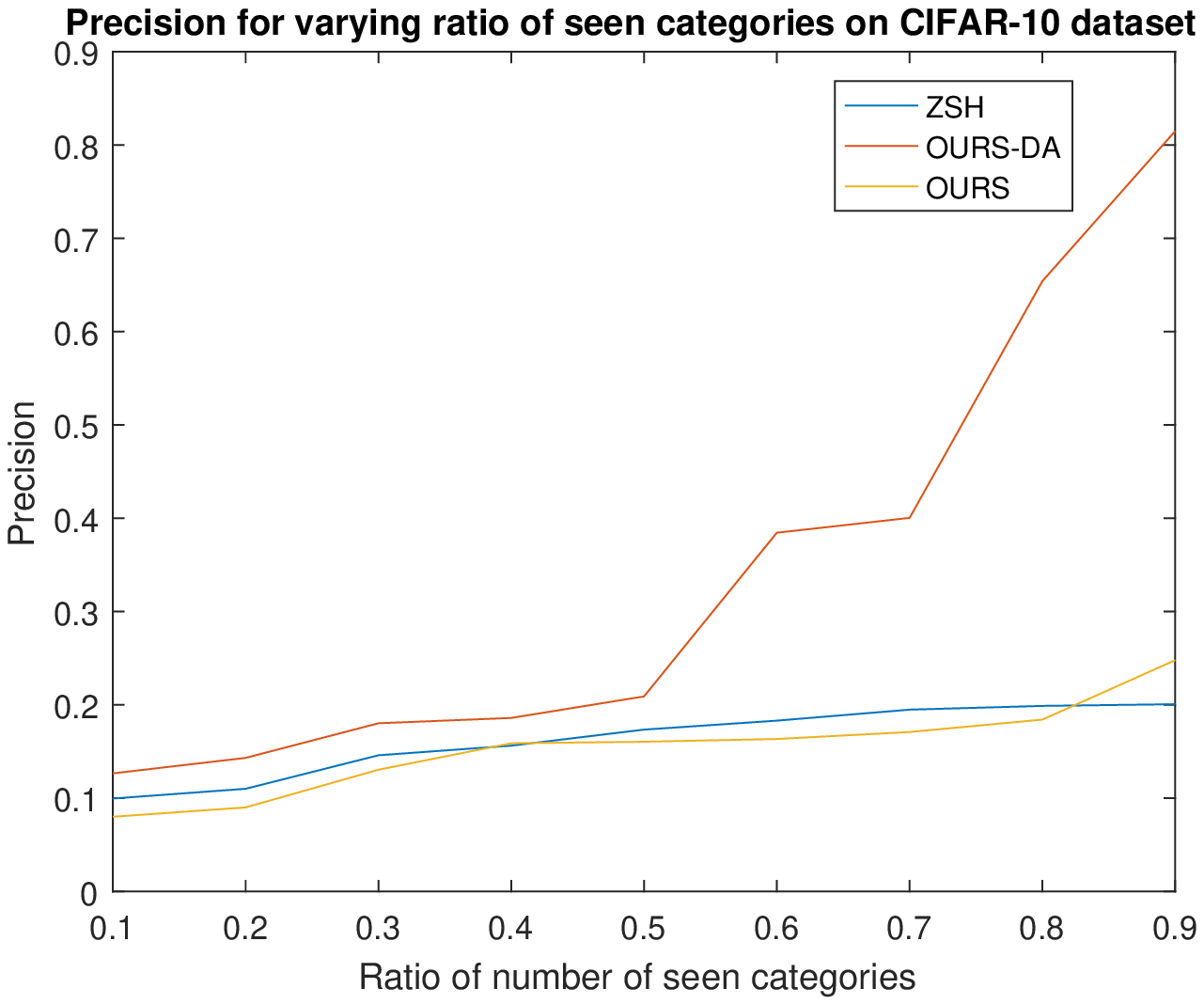}}%
\caption{Effect of the ratio of number of seen categories on the Mean Average Precision (MAP) and Precision metrics on the CUB dataset. \label{EffectofSeenCategoryRatio}}
\end{figure}

In this section, we present the results of the set of experiments performed to evaluate the performance of our proposed algorithm with respect to different numbers of seen categories and training size. Specifically, to observe the number of effect of seen categories, we varied the ratio of seen categories in the training set from 0.1 to 0.9. We took all the images from seen categories for training. Further, we randomly selected 2,500 images from the unseen set as query set to search in the retrieval dataset which is formed by combining the rest of the images of unseen categories with that of seen categories. We fixed the hash code size as 96 bits for all the experiments. We plot the MAP and Precision curves for our method and compared it with the ZSH algorithm in Fig. \ref{EffectofSeenCategoryRatio}. Precision mainly concentrates on the retrieval accuracy and we reported the results with Hamming radius r $\leq$ 2.
From the experiments, we can observe the performance of the algorithms in terms of both the Precision and MAP metrics increases as the ratio of seen categories increases.  This is because our algorithm learns hash function which could produce hash codes with higher discriminablity as the likelihood to find the relevant supervision for the unseen classes increases.  It is also evident from the Fig. \ref{EffectofSeenCategoryRatio} that our method with domain adaptation outperforms ZSH by a very large margin while our method without exploiting domain adaptation has a similar performance as that of ZSH. 

\subsection{Effect of Training Size on CIFAR-10}

\begin{figure}
\centering
\subfigure{\includegraphics[width= 0.5\textwidth]{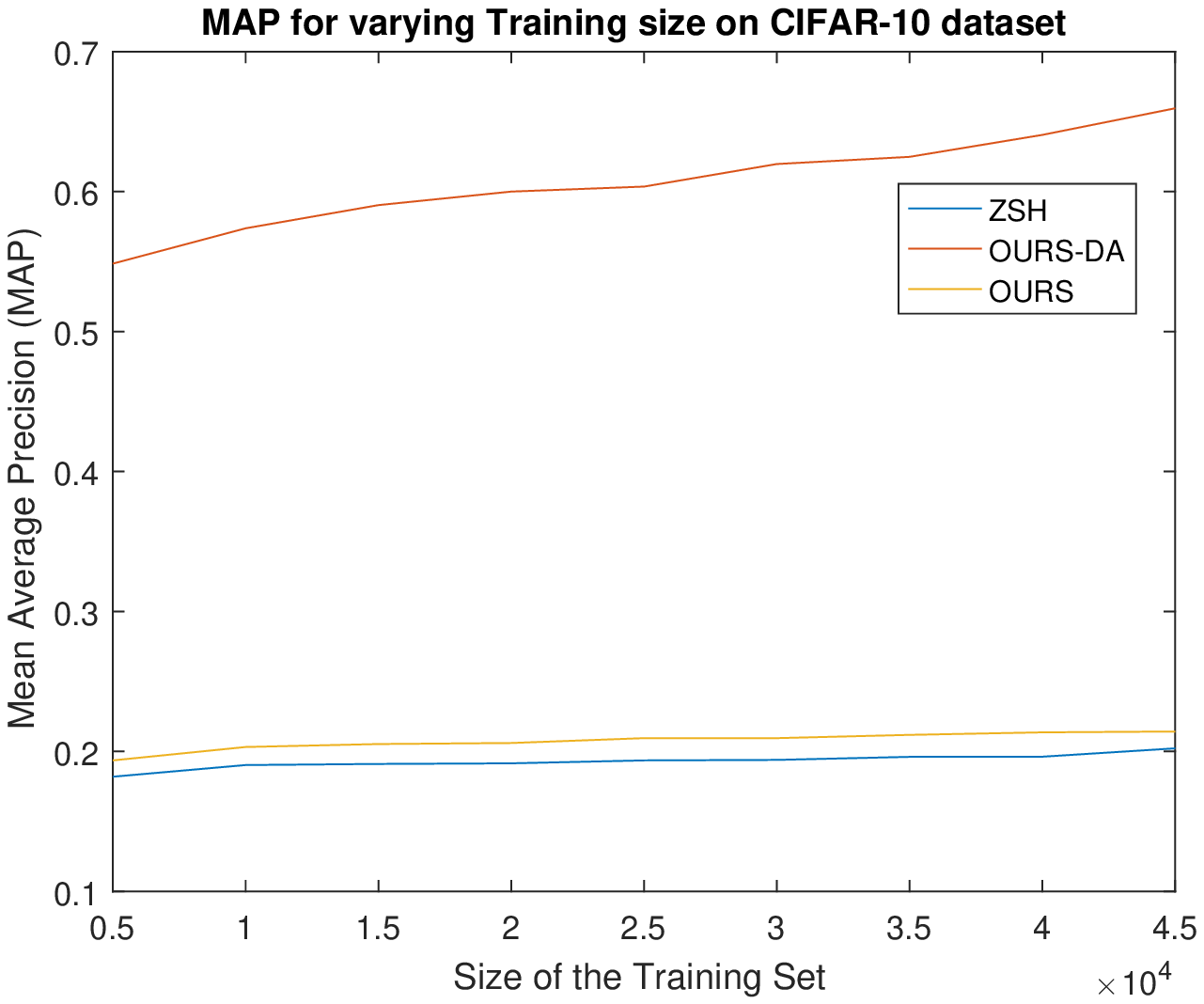}}%
\subfigure{\includegraphics[width= 0.5\textwidth]{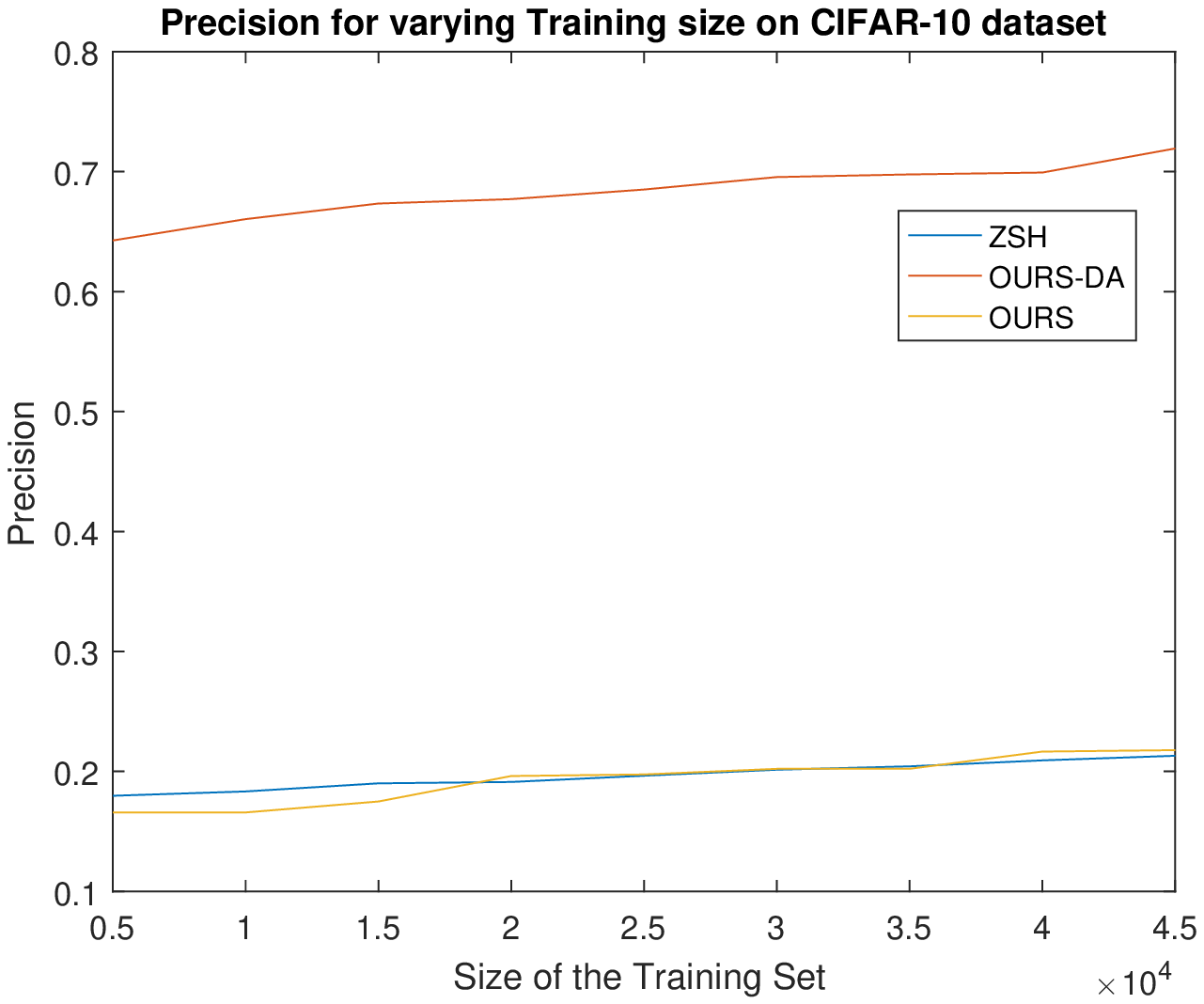}}%
\caption{Effect of the size of the Training dataset on the Mean Average Precision (MAP) and Precision metrics on the CIFAR-10 dataset. \label{EffectofTrainingSize}}
\end{figure}

In this section, we present the results of the set of experiments performed to evaluate the performance of the proposed algorithm with respect to number of training samples from seen categories. We took images from 8 classes as our seen images and images from the rest 2 classes as unseen images. We plot the MAP and Precision curves for our method in Fig. \ref{EffectofTrainingSize} and compare it with the ZSH algorithm by varying the number of training samples from seen classes from 5,000 to 45,000 in the steps of 5,000 samples. At each step, we ensured that we have equal number of samples from each of the seen classes. Further, we randomly selected 2,500 images from the unseen set as query set to search in the retrieval dataset which is formed by combining the rest of the images of unseen categories with all the images (i.e. 48,000 images) of seen categories. We fixed the hash code size as 96 bits for all the experiments. 
From the experiments, we can observe that performance of our algorithm in terms of MAP metric continuously increases as the size of the training set increases from 5,000 to 25,000 samples and then saturates, whereas the Precision metric remains almost the same even with increasing the size of training dataset. As evident from the Fig. \ref{EffectofTrainingSize}, our method without exploiting domain adaptation has a similar performance as that of ZSH while our method integrated with domain adaptation outperforms ZSH by a very large margin. 

\section{Conclusion}

With the escalation of new concepts online and due to the high cost of manual annotations, existing supervised hashing algorithms are unable to report good results on the problem of image retrieval with high accuracy due to the lack of sufficient amount of annotated data. To challenge the issue, in this paper we have proposed a new model to learn an effective hashing function using images and semantic information from limited seen classes. By connecting the dots between the semantic concepts of classes and visual features of images belonging to that class in the common Hamming subspace and using the concepts of max-margin learning, we have learnt a hash function that gives superior results for the task of image retrieval as compared to the existing supervised hashing algorithms as evident from our experiments on three image datasets. The promising performance on the three real-life image datasets demonstrates the potential of our algorithm integrated with domain adaptation model in indexing and searching real-life image data. Instead of performing a naive supervised knowledge transfer from seen categories to unseen categories, thee proposed unsupervised domain adaptation model which learns to embed data in a visual space to semantic space, exploiting the information from seen and unseen categories. In future, to further improve the results, we would like to explore the deep architecture models which have given superior results in many computer vision tasks, instead of using linear classifier. 

\section*{Appendix 1}
In this section, we present the derivation of the Eq. \ref{spectralhashing} from the Eq. \ref{finalobjectivefunction2}. Our main objective is formulated as:
\begin{equation} \label{finalobjectivefunc}
\begin{split}
\underset{\boldsymbol{B},\boldsymbol{W_{txt}}}{\mathrm{\textit{arg} min}} \hspace{0.05cm} ||\boldsymbol{Y^{T}}\boldsymbol{W_{txt}} - \boldsymbol{B}||^{2} + \beta ||\boldsymbol{W_{txt}}||^{2} + \gamma tr(\boldsymbol{B^{T}}\boldsymbol{L}\boldsymbol{B})  \hspace{1.3cm} \\            
\end{split}
\end{equation}
$||\boldsymbol{Y^{T}}\boldsymbol{W_{txt}} - \boldsymbol{B}||^{2} + \beta ||\boldsymbol{W_{txt}}||^{2}$ could be re-written as:
\begin{equation}
\begin{split}
= ||\boldsymbol{Y^{T}}\boldsymbol{W_{txt}} - \boldsymbol{B}||^{2} + \beta ||\boldsymbol{W_{txt}}||^{2}  \hspace{10.73cm} \\ 
= tr\Big( (\boldsymbol{Y^{T}}\boldsymbol{W_{txt}} - \boldsymbol{B})^{T}(\boldsymbol{Y^{T}}\boldsymbol{W_{txt}} - \boldsymbol{B})\Big) + \beta tr\Big(\boldsymbol{W_{txt}^{T}}\boldsymbol{W_{txt}}\Big) \hspace{6.68cm} \\
= tr\Big( \boldsymbol{B^{T}}(\boldsymbol{Y^{T}MY - I})^{T}(\boldsymbol{Y^{T}MY} - \boldsymbol{I})\boldsymbol{B} \Big) + \beta tr\Big(\boldsymbol{W_{txt}^{T}}\boldsymbol{W_{txt}}\Big) \hspace{6.08cm} \\  
= tr\Big( \boldsymbol{B^{T}}(\boldsymbol{Y^{T}MYY^{T}MY} - \boldsymbol{2Y^{T}MY} + \boldsymbol{I}) \boldsymbol{B}\Big) + \beta tr\Big(\boldsymbol{ B^{T}Y^{T}MMYB}\Big) \hspace{4.3cm} \\
= tr\Big( \boldsymbol{B^{T}}(\boldsymbol{Y^{T}MM^{-1}MY} - \boldsymbol{2Y^{T}MY + I})\boldsymbol{B}\Big) \hspace{8.28cm}  \\
= tr \Big(\boldsymbol{B^{T}}(\boldsymbol{I-Y^{T}MY})\boldsymbol{B}\Big) \hspace{11.56cm}\\
= tr \Big(\boldsymbol{B^{T}}\boldsymbol{OB} \Big) \hspace{13.54cm}
\end{split}
\end{equation}
Therefore, the Eq. \ref{finalobjectivefunc} could be re-written as:
\begin{equation}
\begin{split}
= tr \Big(\boldsymbol{B^{T}}\boldsymbol{O}\boldsymbol{B} \Big) + \gamma tr(\boldsymbol{B^{T}}\boldsymbol{L}\boldsymbol{B})\\
= tr \Big(\boldsymbol{B^{T}}\boldsymbol{C}\boldsymbol{B} \Big) \hspace{2.32cm}
\end{split}
\end{equation}
where, $\boldsymbol{C} = (\boldsymbol{O} + \gamma \boldsymbol{L})$.


\begin{thebibliography}{10}

\bibitem{wang2014hashing}
J.~Wang, H.~T. Shen, J.~Song, and J.~Ji, ``Hashing for similarity search: A
  survey,'' \emph{arXiv preprint arXiv:1408.2927}, 2014.

\bibitem{strecha2012ldahash}
C.~Strecha, A.~Bronstein, M.~Bronstein, and P.~Fua, ``Ldahash: Improved
  matching with smaller descriptors,'' \emph{IEEE Transactions on Pattern
  Analysis and Machine Intelligence}, vol.~34, no.~1, pp. 66--78, 2012.

\bibitem{petrovic2010streaming}
S.~Petrovi{\'c}, M.~Osborne, and V.~Lavrenko, ``Streaming first story detection
  with application to twitter,'' in \emph{Human Language Technologies: The 2010
  Annual Conference of the North American Chapter of the Association for
  Computational Linguistics}.\hskip 1em plus 0.5em minus 0.4em\relax
  Association for Computational Linguistics, 2010, pp. 181--189.

\bibitem{liu2014weakly}
X.~Liu, D.~Tao, M.~Song, Y.~Ruan, C.~Chen, and J.~Bu, ``Weakly supervised
  multiclass video segmentation,'' in \emph{Proceedings of the IEEE Conference
  on Computer Vision and Pattern Recognition}, 2014, pp. 57--64.

\bibitem{turian2010word}
J.~Turian, L.~Ratinov, and Y.~Bengio, ``Word representations: a simple and
  general method for semi-supervised learning,'' in \emph{Proceedings of the
  48th annual meeting of the association for computational linguistics}.\hskip
  1em plus 0.5em minus 0.4em\relax Association for Computational Linguistics,
  2010, pp. 384--394.

\bibitem{yang2016zero}
Y.~Yang, Y.~Luo, W.~Chen, F.~Shen, J.~Shao, and H.~T. Shen, ``Zero-shot hashing
  via transferring supervised knowledge,'' in \emph{Proceedings of the 2016 ACM
  on Multimedia Conference}.\hskip 1em plus 0.5em minus 0.4em\relax ACM, 2016,
  pp. 1286--1295.

\bibitem{pachori2016zero}
S.~Pachori and S.~Raman, ``Zero shot hashing,'' \emph{arXiv preprint
  arXiv:1610.02651}, 2016.

\bibitem{gionis1999similarity}
A.~Gionis, P.~Indyk, R.~Motwani \emph{et~al.}, ``Similarity search in high
  dimensions via hashing,'' in \emph{VLDB}, vol.~99, no.~6, 1999, pp. 518--529.

\bibitem{shen2015supervised}
F.~Shen, C.~Shen, W.~Liu, and H.~Tao~Shen, ``Supervised discrete hashing,'' in
  \emph{Proceedings of the IEEE Conference on Computer Vision and Pattern
  Recognition}, 2015, pp. 37--45.

\bibitem{kang2016column}
W.-C. Kang, W.-J. Li, and Z.-H. Zhou, ``Column sampling based discrete
  supervised hashing,'' in \emph{Thirtieth AAAI Conference on Artificial
  Intelligence}, 2016.

\bibitem{gong2011iterative}
Y.~Gong and S.~Lazebnik, ``Iterative quantization: A procrustean approach to
  learning binary codes,'' in \emph{Computer Vision and Pattern Recognition
  (CVPR), 2011 IEEE Conference on}.\hskip 1em plus 0.5em minus 0.4em\relax
  IEEE, 2011, pp. 817--824.

\bibitem{shen2013inductive}
F.~Shen, C.~Shen, Q.~Shi, A.~Van Den~Hengel, and Z.~Tang, ``Inductive hashing
  on manifolds,'' in \emph{Proceedings of the IEEE Conference on Computer
  Vision and Pattern Recognition}, 2013, pp. 1562--1569.

\bibitem{li2015two}
Y.~Li, R.~Wang, H.~Liu, H.~Jiang, S.~Shan, and X.~Chen, ``Two birds, one stone:
  Jointly learning binary code for large-scale face image retrieval and
  attributes prediction,'' in \emph{Proceedings of the IEEE International
  Conference on Computer Vision}, 2015, pp. 3819--3827.

\bibitem{lampert2014attribute}
C.~H. Lampert, H.~Nickisch, and S.~Harmeling, ``Attribute-based classification
  for zero-shot visual object categorization,'' \emph{IEEE Transactions on
  Pattern Analysis and Machine Intelligence}, vol.~36, no.~3, pp. 453--465,
  2014.

\bibitem{akata2013label}
Z.~Akata, F.~Perronnin, Z.~Harchaoui, and C.~Schmid, ``Label-embedding for
  attribute-based classification,'' in \emph{Proceedings of the IEEE Conference
  on Computer Vision and Pattern Recognition}, 2013, pp. 819--826.

\bibitem{romera2015embarrassingly}
B.~Romera-Paredes and P.~Torr, ``An embarrassingly simple approach to zero-shot
  learning,'' in \emph{Proceedings of The 32nd International Conference on
  Machine Learning}, 2015, pp. 2152--2161.

\bibitem{xian2016latent}
Y.~Xian, Z.~Akata, G.~Sharma, Q.~Nguyen, M.~Hein, and B.~Schiele, ``Latent
  embeddings for zero-shot classification,'' in \emph{Proceedings of the IEEE
  Conference on Computer Vision and Pattern Recognition}, 2016, pp. 69--77.

\bibitem{farhadi2009describing}
A.~Farhadi, I.~Endres, D.~Hoiem, and D.~Forsyth, ``Describing objects by their
  attributes,'' in \emph{Computer Vision and Pattern Recognition, 2009. CVPR
  2009. IEEE Conference on}.\hskip 1em plus 0.5em minus 0.4em\relax IEEE, 2009,
  pp. 1778--1785.

\bibitem{socher2013zero}
R.~Socher, M.~Ganjoo, C.~D. Manning, and A.~Ng, ``Zero-shot learning through
  cross-modal transfer,'' in \emph{Advances in neural information processing
  systems}, 2013, pp. 935--943.

\bibitem{frome2013devise}
A.~Frome, G.~S. Corrado, J.~Shlens, S.~Bengio, J.~Dean, T.~Mikolov
  \emph{et~al.}, ``Devise: A deep visual-semantic embedding model,'' in
  \emph{Advances in neural information processing systems}, 2013, pp.
  2121--2129.

\bibitem{oliva2001modeling}
A.~Oliva and A.~Torralba, ``Modeling the shape of the scene: A holistic
  representation of the spatial envelope,'' \emph{International journal of
  computer vision}, vol.~42, no.~3, pp. 145--175, 2001.

\bibitem{mikolov2013distributed}
T.~Mikolov, I.~Sutskever, K.~Chen, G.~S. Corrado, and J.~Dean, ``Distributed
  representations of words and phrases and their compositionality,'' in
  \emph{Advances in neural information processing systems}, 2013, pp.
  3111--3119.

\bibitem{pennington2014glove}
J.~Pennington, R.~Socher, and C.~D. Manning, ``Glove: Global vectors for word
  representation.'' in \emph{EMNLP}, vol.~14, 2014, pp. 1532--43.

\bibitem{kodirov2015unsupervised}
E.~Kodirov, T.~Xiang, Z.~Fu, and S.~Gong, ``Unsupervised domain adaptation for
  zero-shot learning,'' in \emph{Proceedings of the IEEE International
  Conference on Computer Vision}, 2015, pp. 2452--2460.

\bibitem{weiss2009spectral}
Y.~Weiss, A.~Torralba, and R.~Fergus, ``Spectral hashing,'' in \emph{Advances
  in neural information processing systems}, 2009, pp. 1753--1760.

\bibitem{shen2015learning}
F.~Shen, W.~Liu, S.~Zhang, Y.~Yang, and H.~T. Shen, ``Learning binary codes for
  maximum inner product search,'' in \emph{2015 IEEE International Conference
  on Computer Vision (ICCV)}.\hskip 1em plus 0.5em minus 0.4em\relax IEEE,
  2015, pp. 4148--4156.

\bibitem{rastegari2012attribute}
M.~Rastegari, A.~Farhadi, and D.~Forsyth, ``Attribute discovery via predictable
  discriminative binary codes,'' in \emph{European Conference on Computer
  Vision}.\hskip 1em plus 0.5em minus 0.4em\relax Springer, 2012, pp. 876--889.

\bibitem{fan2008liblinear}
R.-E. Fan, K.-W. Chang, C.-J. Hsieh, X.-R. Wang, and C.-J. Lin, ``Liblinear: A
  library for large linear classification,'' \emph{Journal of machine learning
  research}, vol.~9, no. Aug, pp. 1871--1874, 2008.

\bibitem{fu2014transductive}
Y.~Fu, T.~M. Hospedales, T.~Xiang, Z.~Fu, and S.~Gong, ``Transductive
  multi-view embedding for zero-shot recognition and annotation,'' in
  \emph{European Conference on Computer Vision}.\hskip 1em plus 0.5em minus
  0.4em\relax Springer, 2014, pp. 584--599.

\bibitem{margolis2011literature}
A.~Margolis, ``A literature review of domain adaptation with unlabeled data,''
  \emph{Tec. Report}, pp. 1--42, 2011.

\bibitem{gan2016learning}
C.~Gan, T.~Yang, and B.~Gong, ``Learning attributes equals multi-source domain
  generalization,'' in \emph{Proceedings of the IEEE Conference on Computer
  Vision and Pattern Recognition}, 2016, pp. 87--97.

\bibitem{krizhevsky2009learning}
A.~Krizhevsky, ``Learning multiple layers of features from tiny images,'' 2009.

\bibitem{WahCUB_200_2011}
C.~Wah, S.~Branson, P.~Welinder, P.~Perona, and S.~Belongie, ``The caltech-ucsd
  birds-200-2011 dataset,'' California Institute of Technology, Tech. Rep.
  CNS-TR-2011-001, 2011.

\bibitem{lampert2009learning}
C.~H. Lampert, H.~Nickisch, and S.~Harmeling, ``Learning to detect unseen
  object classes by between-class attribute transfer,'' in \emph{Computer
  Vision and Pattern Recognition, 2009. CVPR 2009. IEEE Conference on}.\hskip
  1em plus 0.5em minus 0.4em\relax IEEE, 2009, pp. 951--958.

\bibitem{turpin2006user}
A.~Turpin and F.~Scholer, ``User performance versus precision measures for
  simple search tasks,'' in \emph{Proceedings of the 29th annual international
  ACM SIGIR conference on Research and development in information
  retrieval}.\hskip 1em plus 0.5em minus 0.4em\relax ACM, 2006, pp. 11--18.

\bibitem{huang2012improving}
E.~H. Huang, R.~Socher, C.~D. Manning, and A.~Y. Ng, ``Improving word
  representations via global context and multiple word prototypes,'' in
  \emph{Proceedings of the 50th Annual Meeting of the Association for
  Computational Linguistics: Long Papers-Volume 1}.\hskip 1em plus 0.5em minus
  0.4em\relax Association for Computational Linguistics, 2012, pp. 873--882.

\bibitem{simonyan2014very}
K.~Simonyan and A.~Zisserman, ``Very deep convolutional networks for
  large-scale image recognition,'' \emph{arXiv preprint arXiv:1409.1556}, 2014.

\bibitem{vedaldi2015matconvnet}
A.~Vedaldi and K.~Lenc, ``Matconvnet: Convolutional neural networks for
  matlab,'' in \emph{Proceedings of the 23rd ACM international conference on
  Multimedia}.\hskip 1em plus 0.5em minus 0.4em\relax ACM, 2015, pp. 689--692.

\bibitem{donahue2014decaf}
J.~Donahue, Y.~Jia, O.~Vinyals, J.~Hoffman, N.~Zhang, E.~Tzeng, and T.~Darrell,
  ``Decaf: A deep convolutional activation feature for generic visual
  recognition.'' in \emph{ICML}, 2014, pp. 647--655.

\end{thebibliography}
\end{document}